\documentclass{article}

\usepackage{microtype}
\usepackage{graphicx}
\usepackage{subcaption}
\usepackage{booktabs} 
\usepackage{multirow}
\usepackage{hyperref}

\usepackage[accepted]{icml2026}

\usepackage{amsmath}
\usepackage{amssymb}
\usepackage{mathtools}
\usepackage{amsthm}
\usepackage{enumitem}
\usepackage{xcolor}
\usepackage{hyperref}

\usepackage[capitalize,noabbrev]{cleveref}

\theoremstyle{plain}

\theoremstyle{definition}

\theoremstyle{remark}

\usepackage[textsize=tiny]{todonotes}

\icmltitlerunning{Evolving Unified Multimodal Models into Self-Adaptive Interleaved Visual Reasoners}

\begin{document}

\twocolumn[
  \icmltitle{Breaking Dual Bottlenecks: Evolving Unified Multimodal Models \\ into Self-Adaptive Interleaved Visual Reasoners}



  \icmlsetsymbol{equal}{*}

  \begin{icmlauthorlist}
    \icmlauthor{Qingyang Liu}{yyy,comp}
    \icmlauthor{Bingjie Gao}{yyy}
    \icmlauthor{Canmiao Fu}{comp}
    \icmlauthor{Zhipeng Huang}{comp}
    \icmlauthor{Chen Li}{comp}
    \icmlauthor{Feng Wang}{comp}
    \icmlauthor{Shuochen Chang}{yyy}
    \icmlauthor{Shaobo Wang}{yyy}
    \icmlauthor{Yali Wang}{CAS}
    \icmlauthor{Keming Ye}{comp}
    \icmlauthor{Jiangtong Li}{TJU}
    \icmlauthor{Li Niu}{yyy,MG}
  \end{icmlauthorlist}

  \icmlaffiliation{yyy}{Shanghai Jiao Tong University}
  \icmlaffiliation{comp}{WeChat Vision, Tencent Inc.}
  \icmlaffiliation{CAS}{Shenzhen Institutes of Advanced Technology, Chinese Academy of Sciences}
  \icmlaffiliation{TJU}{Tongji University}
  \icmlaffiliation{MG}{miguo.ai}

  \icmlcorrespondingauthor{Qingyang Liu}{narumimaria@sjtu.edu.cn}
  \icmlcorrespondingauthor{Jiangtong Li}{jiangtongli@tongji.edu.cn}
  \icmlcorrespondingauthor{Li Niu}{ustcnewly@sjtu.edu.cn}

  \icmlkeywords{Machine Learning, ICML}

  \vskip 0.3in
]



\printAffiliationsAndNotice{}  

\begin{abstract}
  Recent unified models integrate multimodal understanding and generation within a single framework.
  However, an ``understanding-generation gap'' persists, where models can capture user intent but often fail to translate this semantic knowledge into precise pixel-level manipulation.
  This gap results in two bottlenecks in anything-to-image task~(X2I): the \textbf{attention entanglement bottleneck}, where blind planning struggles with complex prompts, and the \textbf{visual refinement bottleneck}, where unstructured feedback fails to correct imperfections efficiently.
  In this paper, we propose a novel framework that empowers unified models to autonomously switch between generation strategies based on instruction complexity and model capability.
  To achieve this, we construct a hierarchical data pipeline that constructs execution paths across three adaptive modes: direct generation for simple cases, self-reflection for quality refinement, and multi-step planning for decomposing complex scenarios.
  Building on this pipeline, we contribute a high-quality dataset with over 50,000 samples and implement a two-stage training strategy comprising SFT and RL.
  Specifically, we design step-wise reasoning rewards to ensure logical consistency and intra-group complexity penalty to prevent redundant computational overhead.
  Extensive experiments demonstrate that our method outperforms existing baselines on X2I, achieving superior generation fidelity among simple-to-complex instructions. The code is released at \href{https://github.com/WeChatCV/Interleaved_Visual_Reasoner}{GitHub}.
\end{abstract}

\section{Introduction}
\begin{figure*}[h]
  \centering
  \includegraphics[width=0.98\textwidth]{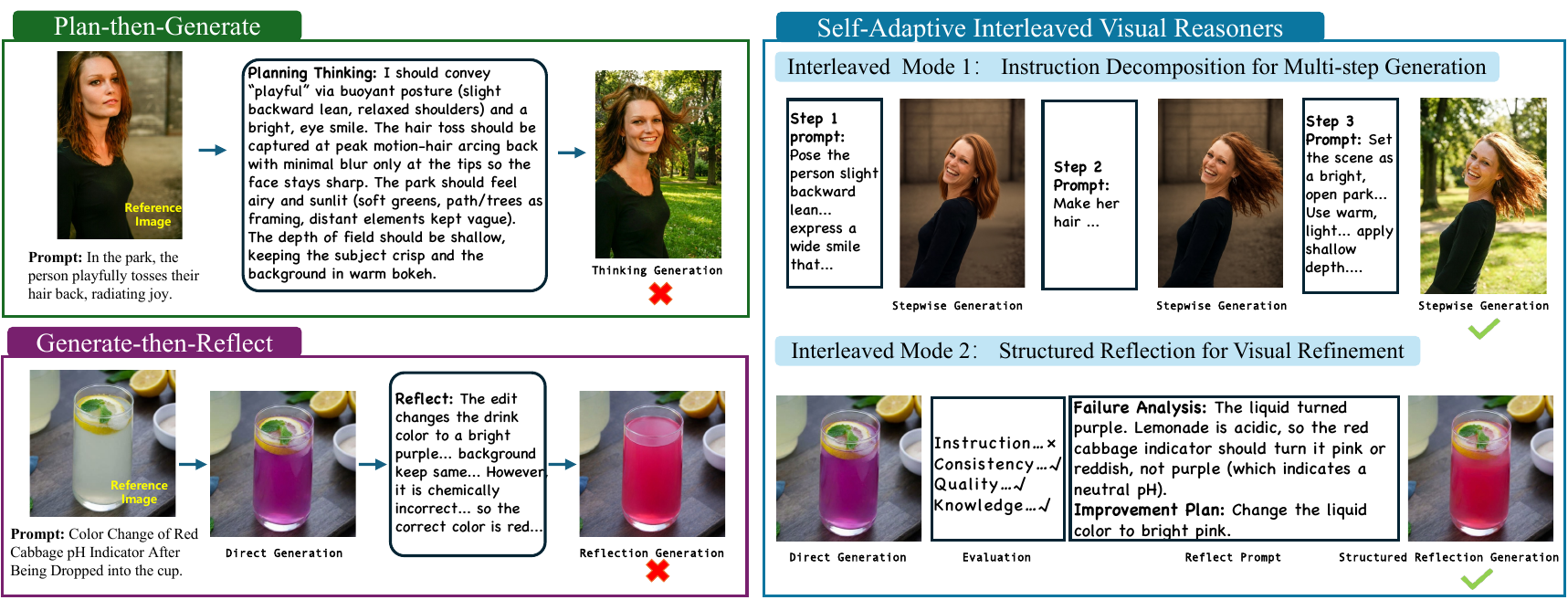}
  \caption{Comparison of conventional paradigms and our Self-Adaptive Interleaved Visual Reasoners. 
    \textbf{Left:} Existing methods struggle with ``blind'' planning and unstructured reflection due to understanding-generation gap. 
    \textbf{Right:} Our approach adopts adaptive strategies: \textbf{Mode 1} decomposes complex prompts via interleaved generation; \textbf{Mode 2} employs structured reflection for interleaved visual refinement.}
\label{fig:comparison}
  \label{fig:task}
  \vskip -1.1em
\end{figure*}


Recent advancements in unified large multimodal models~(unified models)~\cite{wu2025janus, wang2024emu3} have integrated multimodal understanding and generation within a single framework.
Inspired by the success of Chain-of-Thought (CoT) reasoning, recent studies~\cite{ye2025visual, zhao2025learning, han2026unicorn} have incorporated multimodal CoT reasoning into these unified architectures.
This integration enables complex anything-to-image (X2I) tasks~\cite{liu2025step1x, niu2025wise, liu2024shadow, zhao2025shadow,zhao2026texeditor}, requiring models to generate or edit images conditioned on diverse multimodal inputs (\emph{e.g.}, text, layouts, sketches).
Although these models demonstrate promising visual reasoning and generative capabilities, a critical disparity remains between their understanding potential and actual generation fidelity~\cite{chen2025empirical}.
Specifically, while current unified models exhibit strong visual comprehension, their ability to execute precise, instruction-following image editing often lags behind.
We term this issue the ``\textit{\textbf{understanding-generation gap}}'', which becomes evident in complex X2I scenarios: the model ``\textit{\textbf{knows}}'' the user's intent but fails to ``\textit{\textbf{faithfully translate}}'' this semantic knowledge into accurate pixel-level manipulation.


Driven by this understanding-generation gap, unified models face two primary challenges when executing complex tasks: the {attention entanglement bottleneck~\cite{park2025fair, koh2025translation}} and the visual refinement bottleneck~\cite{wu2025omnigen2}.
The attention entanglement bottleneck stems from the complexity of user prompts, which act as a barrier to direct, single-pass synthesis and require decomposing instructions into step-by-step edits.
Meanwhile, the visual refinement bottleneck arises from unavoidable imperfections in single-step pixel synthesis, creating a need for progressive visual refinement.
To mitigate the attention entanglement bottleneck, recent approaches formulate detailed textual plans before execution~\cite{qin2025uni, ye2025loom}.
However, these methods lack verification mechanisms; the reasoning process remains ``blind'' to the model's actual generative limits, often yielding unexecutable plans that fail to handle complex, multi-step editing.
To handle the visual refinement bottleneck, several methods employ critiques of initial outputs to guide subsequent refinement~\cite{guo2025can,zhuo2025reflection}.
These methods typically rely on switching between multiple distinct models, which incurs high computational costs and obstructs automated, progressive self-improvement.
Recent efforts attempt to iteratively generate and critique images using a unified model within a feedback loop~\cite{guo2025thinking}.
However, such unstructured critiques mix error analysis and improvement suggestion together, making it difficult to resolve compound errors efficiently.
Crucially, no existing work has successfully targeted both bottlenecks simultaneously within a unified framework.


To overcome these two bottlenecks, we propose a Self-adaptive Interleaved Reasoner, which equips the unified models with native interleaved generation and dynamic self-evaluation capabilities, allowing it to autonomously switch strategies based on instruction complexity and model capability.
Fig.~\ref{fig:task} illustrates two interleaved reasoning modes of our reasoner.
Specifically, for complex multi-intent instructions, the reasoner utilizes textual planning and interleaved generation to decompose tasks step-by-step.
When initial outputs fail to meet quality standards, the reasoner switches to a self-reflection mode for progressive visual refinement.
To develop this adaptive capability, we propose a data pipeline driven by a hierarchical escalation mechanism.
Given a raw X2I~\cite{qin2023gluegen} input, we first employ the baseline unified models to directly generate the image.
We then use an \textsc{Analyzer}, Qwen3vl-235B~\cite{Qwen3-VL}, as an independent critic to evaluate the output's alignment with the user instruction and reference inputs.
If the initial generation is unsatisfactory, the \textsc{Analyzer} diagnoses specific errors in the faulty image and synthesizes a reflection prompt.
Guided by this reflection, a \textsc{Generator}, Gemini-3-Pro-Image~\cite{google2025gemini3proimage}, attempts to rectify the output.
If the task remains unresolved after the maximum iteration limit, the \textsc{Analyzer} identifies reason for such failure. 
Cases identified as having excessive prompt complexity escalate to the multi-step planning phase, while others (\emph{e.g.}, those requiring specialized knowledge) are filtered out.
For multi-step planning phase, the LMM decomposes the complex instruction into sequential sub-tasks, enabling step-by-step generation and intermediate evaluation until the objective is met.
Following strict human verification, we obtain a high-fidelity interleaved dataset with over 50,000 samples that integrates generation, reflection, planning, and evaluation across three adaptive modes.

Leveraging this constructed dataset, we implement a two-stage training paradigm.
First, we apply Supervised Fine-Tuning (SFT) to adapt the model to the interleaved reasoning syntax and basic instruction following. Subsequently, we employ Reinforcement Learning~(RL) with Group Relative Policy Optimization algorithm (GRPO) to optimize the model's strategic planning, ensuring it selects the most effective generation mode for varying instruction complexities.
Beyond standard format and outcome rewards assessed by an LMM, we introduce a step-wise reasoning reward and an intra-group complexity penalty.
The step-wise reasoning reward verifies the logical consistency of intermediate steps, utilizing the LMM to validate each textual output.
To prevent inefficient over-reasoning, the intra-group complexity penalty favors simpler generation modes when they yield results equivalent to complex ones, thereby balancing generation quality with computational efficiency.
Extensive experiments and visualization results confirm that our method outperforms existing baselines~(EUM3.5~\cite{cui2025emu3}, \emph{etc.}) among text-to-image, image editing, and X2I tasks.
Our contributions can be summarized as:

\begin{itemize}[topsep=0pt]
\setlength{\itemsep}{0pt}
\setlength{\parsep}{0pt}
\setlength{\parskip}{3pt}
    \item We propose a unified framework with interleaved plan or reflect, resolving the understanding-generation gap of unified models by addressing attention entanglement and visual refinement bottlenecks.
    \item We design a hierarchical data synthesis pipeline that constructs execution paths across direct, reflective, and multi-step planning modes. Based on this pipeline, we contribute a large-scale high-fidelity dataset to enable adaptive X2I generation.
    \item We implement a two-stage training strategy with SFT and RL. With our proposed step-wise reasoning reward and intra-group complexity penalty, we ensure logical correctness while minimizing computational costs.
\end{itemize}

\section{Related Work}
\subsection{Unified Large Multimodal Models}
The field of unified multimodal understanding and generation~\cite{pan2025transfer, shi2024lmfusion, sun2023emu, ge2024seed, liao2025divide, chang2026d3tom} has advanced rapidly in recent years.
Current unified models generally fall into three categories: diffusion-based~\cite{swerdlow2025unified, li2025dual, yang2505mmada}, autoregressive~\cite{wu2025harmonizing, li2025onecat}, and hybrid autoregressive–diffusion architectures~\cite{xie2024show, zhou2024transfusion, ma2025janusflow}.
Diffusion-based models generate images by iterative denoising, while autoregressive models predict visual and textual tokens sequentially. 
Hybrid architectures combine autoregressive reasoning with diffusion decoding for structured planning and high-quality synthesis. 
However, unified models still suffer from an understanding--generation gap, where semantic understanding does not reliably translate into pixel-level outputs. 
This gap creates two bottlenecks in X2I tasks: the \textbf{attention entanglement bottleneck}, where blind planning fails on complex prompts, and the \textbf{visual refinement bottleneck}, where unstructured feedback cannot efficiently correct errors. 
Thus, a unified solution is needed to tightly couple reasoning and generation for adaptive instruction decomposition and output refinement.

\subsection{Reasoning in Generation}
Recent researches increasingly focus on the reasoning and reflection capabilities of LMMs~\cite{wang2024qwen2, hurst2024gpt, liu2026bridging, peng2026hyperet, ma2025human, han2026dcr}.
Beyond their success in multimodal comprehension, studies now apply these capabilities to enhance image/video generation and editing~\cite{han2025lightfair, fang2025got, huang2025interleaving, jiang2025draco, liu2026animatescene, chen2025tivibench, ye2025unicedit, wu2025cinetrans, han2024aucseg,wang2025sega,zhao2026making}.
LMMs are employed to analyze intermediate results, identifying inconsistencies or instruction violations to guide subsequent generation steps.
These reasoning-driven pipelines allow models to refine outputs through self-evaluation, improving instruction adherence and visual fidelity in complex scenarios.
Recently, this strategy has been integrated into unified models~\cite{yin2025reasonedit}, which jointly support understanding and generation within a single framework.
By aligning semantic reasoning with generation in a unified framework, unified models incorporate planning and corrective feedback directly into the generation process. 
Many reasoning-in-generation methods use RL~\cite{shao2024deepseekmath, rafailov2023direct} with self-evaluation or external feedback as rewards, making reasoning trainable for better instruction following and iterative correction. 
However, existing approaches still fail to close the understanding--generation gap, as they mostly follow rigid ``Plan-then-Generate''~\cite{jiang2025t2i, liao2025imagegen, gao2025devil} or ``Generate-then-Reflect''~\cite{li2025reflect, gao2025rapo++} pipelines. 
Lacking adaptive control over when to plan or reflect, they often incur blind planning or inefficient feedback, and thus cannot simultaneously resolve attention entanglement and visual refinement bottlenecks in complex X2I tasks.

\begin{figure*}[h]
  \centering
  \includegraphics[width=\textwidth]{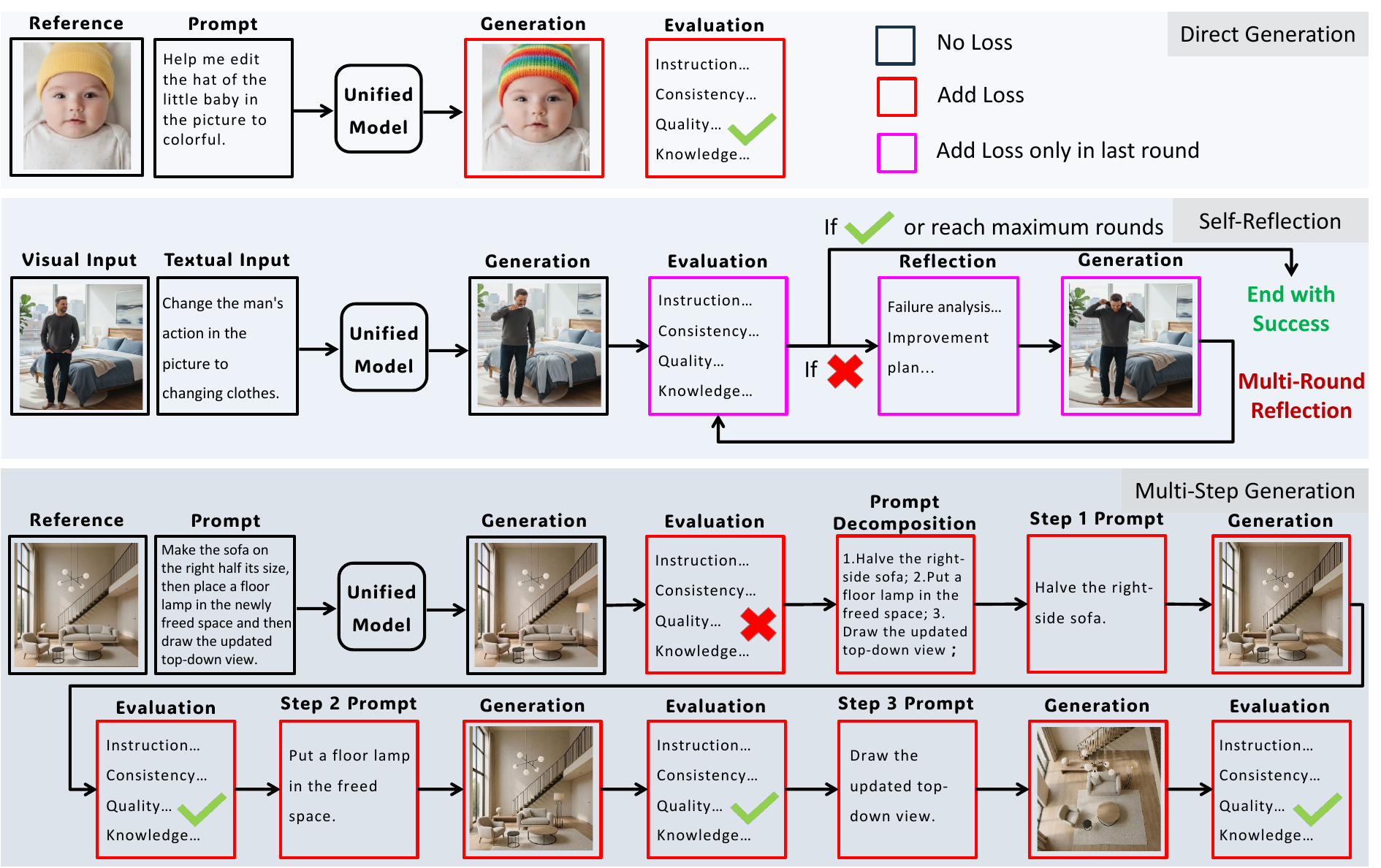}
  \caption{Illustration of the three kinds of training data in our dataset and the selective loss masking strategy for in the SFT stage.}
  \label{fig:data}
  \vskip -1em
\end{figure*}

\section{Data Construction}
\label{sec:data}
To equip the unified model with broad generalization capabilities, we construct a dataset comprising diverse source modalities paired with user instructions of varying complexity.
To enable the unified model to switch between generation, reflection, and planning, we develop an automated, hierarchical data construction pipeline.
This pipeline employs the \textsc{Analyzer} (Qwen-235B) as an evaluator, failure diagnostician, and task planner, alongside the \textsc{Generator} (Gemini-3-Pro-Image), which synthesizes images based on reflection prompts or step-wise instructions.
The construction process automatically categorizes data into three operational modes according to instruction complexity.
Fig.~\ref{fig:data} illustrates representative cases from each mode, demonstrating the effectiveness of our data construction pipeline.

\subsection{Direct Generation}
We initiate the process by prompting the baseline unified model for direct generation, after which the \textsc{Analyzer} evaluates the output across four criteria:
\textbf{1. Instruction:} Verifies whether the user instructions are accurately executed.
\textbf{2. Consistency:} Checks if the identity and attributes of unedited regions in the reference image remain preserved.
\textbf{3. Quality:} Assesses overall visual fidelity, identifying potential artifacts or degradation.
\textbf{4. Knowledge:} Confirms that the generated content aligns with physical laws and commonsense.
If the output meets all criteria, we retain the trajectory as a direct generation instance, representing scenarios where single-step inference suffices.
For the direct generation mode, each data sample consists of the generated image $G_{1}$ and its corresponding evaluation $E_{1}$.

\subsection{Self-Reflection}
If the initial generation fails validation, the pipeline initiates a self-correction loop, limited to a maximum of three iterations.
The \textsc{Analyzer} processes the original instruction, the reference image, the evaluation text and the rejected output to formulate a targeted ``reflection prompt.''
This prompt consists of two key components: a failure analysis that identifies specific errors, and an improvement plan that guides the subsequent generation.
Guided by this reflection, the \textsc{Generator} attempts to correct the image errors.
Once a revised output meets the \textsc{Analyzer}'s evaluation standards, the iterative loop terminates.
We then record the successful trajectory as a self-reflection data instance, indicating that the self-correction is sufficient to resolve the failure.
Each data sample for self-reflection is defined as $\bigcup_{i=1}^{K-1} \{G_{i}, E_{i}, R_{i}\} \cup\{G_{K}, E_{K}\}$, where $G_{i}, E_{i}$, and $R_{i}$ denote the generated image, evaluation, and reflection at the $i$-th iteration, respectively, and $K$ represents the total number of iterations.


\subsection{Multi-step Generation}
If the task remains unsolved after reaching the maximum three-iteration limit, the pipeline avoids automatic escalation.
Instead, the \textsc{Analyzer} performs a comprehensive diagnosis to identify the root cause of the failure.
Specifically, if the failure stems from excessive prompt complexity, the pipeline escalates to explicit task decomposition; otherwise, cases attributed to other factors (\emph{e.g.}, lack of domain knowledge) are filtered out.
In this phase, the \textsc{Analyzer} acts as a planner to break down the original instruction into sequential sub-tasks, which the \textsc{Generator} executes one by one while the \textsc{Analyzer} evaluates each intermediate step result.
Upon successful completion of the sequence, we perform trajectory pruning, filtering out the previously failed reflection attempts.
This process yields a clean sequential reasoning path, constituting the Multi-step Generation data.
We archive this pruned trajectory as a multi-step instance, representing the explicit sequential task decomposition are necessary to resolve complex instructions.
Each data sample for multi-step generation is defined as $\{ G_1, E_1 \} \cup \bigcup_{i=2}^{N+1} \{ S_i, G_i, E_i \}$, where, $G_{1}$ is the direct generation and $E_{1}$ is its failure analysis; $S_i, G_i$, and $E_i$ represent the sub-instruction, generated image, and evaluation at sub-step $i-1$, with $N$ denoting the number of planning steps for the multi-step generation.

\begin{figure*}[h]
  \centering
  \includegraphics[width=\textwidth]{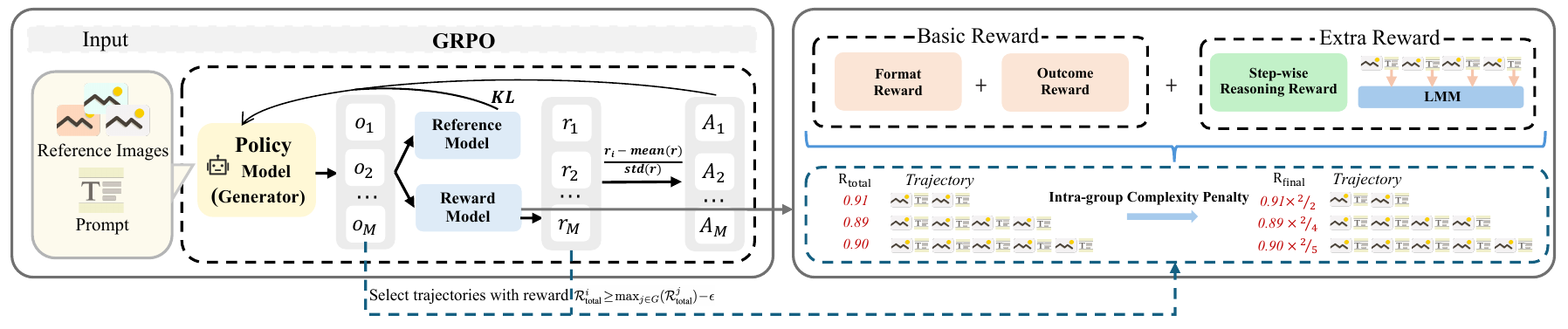}
  \caption{Overview of the GRPO-based RL stage.
\textbf{Left:} The Policy Model generates a group of candidate trajectories ($o_1 \dots o_M$) to compute group-relative advantages ($A_i$).
\textbf{Right:} The composite reward function aggregates \textbf{Basic Rewards} (Format, Outcome) and an \textbf{Extra Reward} (Step-wise Reasoning) validated by an LMM.
An \textbf{Intra-group Complexity Penalty} modulates the Total Reward ($R_{\text{total}}$): among successful trajectories, those with redundant steps are scaled down to derive the Final Reward ($R_{\text{final}}$), favoring effective solution.}
  \label{fig:rl}
  \vskip -1em
\end{figure*}

\subsection{Human Verification}
To guarantee the high quality of our hybrid dataset, every synthesized instance undergoes a strict human verification process.
For each data sample, we employ two human annotators to review both the final generation outcomes and the logical coherence of all intermediate reasoning and planning steps.
Only instances accepted by all annotators are retained as valid cases.
Through this combination of automated construction and manual quality assurance, the resulting dataset enables the model to perform iterative self-reflection and multi-step decomposition, grounding high-fidelity generation in text-based reasoning signals.

\section{Training}
We employ a two-stage training strategy to equip the unified model with adaptive reasoning skills and computational efficiency.
In the first stage, supervised fine-tuning adapts the model to the interleaved format, enabling it to execute textual planning and visual reflection during generation.
In the second stage, we explore reinforcement learning with GRPO to optimize the unified model, training the model to select the most efficient execution path for a given task.

\subsection{Supervised Fine-Tuning}
In the SFT stage, we train the unified model using our curated dataset, which covers all three operational modes.
Standard training on this dataset would require the model to predict suboptimal intermediate images, potentially degrading its generative fidelity.
To prevent this, we employ a selective loss masking strategy. 
As shown in Fig.~\ref{fig:data}, given an instruction-reference pair $c$, the model minimizes the standard auto-regressive negative log-likelihood loss, computed on a selected subset of the output sequence $\mathcal{O}$:
\begin{equation}
  \mathcal{L}_{\text{SFT}} = - \sum_{t \in \mathcal{O}} \log P(x_t | x_{<t}, c),
\end{equation}
where $x_t$ represents the $t$-th token (text or image patch). 
The target subset $\mathcal{O}$ is defined according to the data mode.
\\\textbf{Direct Generation: }For direct execution, the loss is applied to both the generated image $G_{1}$ and its evaluation $E_{1}$, such that $\mathcal{O}_{\text{direct}} = \{ G_{1}, E_{1} \}$.
\\\textbf{Self-Reflection: }Given a reflection trajectory with $k$ attempts (where only the $K$-th succeeds), we ensure the model learns correction logic without memorizing the artifacts of failed generations.
Therefore, we mask the loss for all prior failed generations $\{G_1, ..., G_{K-1}\}$ and their evaluations up to step $K-2$.
The loss is applied only to the final correction phase: the diagnosis of the last failure $E_{K-1}$, the reflection prompt $R_{K-1}$ (containing analysis and planning), and the final successful generation $G_{K}$ with its evaluation $E_{K}$, defined as $\mathcal{O}_{\text{reflect}} = \{ E_{K-1}, R_{K-1}, G_{K}, E_{K} \}$.
\\\textbf{Multi-step Generation:} To learn explicit task decomposition, we compute loss on the evaluation of the initial direct attempt $E_{1}$, followed by the complete multi-step planning sequence. For each sub-step $i-1$, the loss includes the sub-instruction $S_i$, the intermediate generation $G_i$, and its validation $E_i$, given by $\mathcal{O}_{\text{multi}} = \{ E_{1} \} \cup \bigcup_{i=2}^{N+1} \{ S_i, G_i, E_i \}$.
Through this masking strategy, the unified model learns self-reflection and multi-step decomposition while avoiding training on low-quality visual intermediates.

\begin{table*}[t]
    \centering
    \caption{Quantitative evaluation results on the KRIS-Bench benchmark. Best performance are highlighted \textbf{bolded}.}
    \label{tab:krisbench}
    \setlength{\tabcolsep}{11pt} 
    \resizebox{\linewidth}{!}{
    \begin{tabular}{llcccc}
    \toprule
     & \textbf{Models} & \textbf{Factual} & \textbf{Conceptual} & \textbf{Procedural} & \textbf{Overall} \\
    \midrule
    
    \multirow{3}{*}{\textbf{\shortstack{Close-source\\Model}}} 
      & Gemini 2.5 flash & 77.03 & 78.29 & 75.93 & 77.29 \\
      & Doubao  & 78.10 & 76.86 & 76.93 & 77.31 \\
      & GPT4o  & 79.80 & 81.37 & 78.32 & 80.09 \\
    \midrule
    
    \multirow{14}{*}{\textbf{\shortstack{Open-Source\\Model}}} 
      & OmniGen2~\cite{wu2025omnigen2} & 57.36 & 44.20 & 47.79 & 49.71 \\
      & BAGEL-thinking~\cite{deng2025bagel} & 66.18 & 61.92 & 49.02 & 60.18 \\
      & BAGEL~\cite{deng2505emerging} & 60.26 & 55.86 & 51.69 & 56.21 \\
      & Uniworld-V1~\cite{lin2025uniworld} & 47.71 & 44.80 & 47.92 & 50.27 \\
      & Flux-Kontext-dev~\cite{labs2025flux} & 53.28 & 50.36 & 42.53 & 49.54 \\
      & UniWorld-FLUX.1-Kontext-Dev~\cite{li2025uniworld} & 55.50 & 51.39 & 43.76 & 51.04 \\
      & UniWorld-Qwen-Image-Edit~\cite{li2025uniworld} & 61.72 & 56.38 & 46.69 & 55.98 \\
      & Step1X-Edit v1.1~\cite{liu2025step1x} & 53.05 & 54.34 & 44.66 & 51.59 \\
      & Qwen-Image-Edit-2509~\cite{wu2025qwen} & 61.47 & 56.79 & 47.07 & 56.15 \\
      & ReasonEdit-Q (thinking+reflection)~\cite{yin2025reasonedit} & 63.92 & 64.85 & 52.41 & 61.57 \\
    \cmidrule{2-6}
      & Emu3.5~\cite{cui2025emu3} & 78.59 & 71.92 & 71.14 & 73.75 \\
      & \textbf{Ours} & \textbf{84.24} & \textbf{74.83} & \textbf{85.53} & \textbf{80.18} \\
    \bottomrule
    \end{tabular}}
    \vskip -1em
\end{table*}

\subsection{Reinforcement Learning Stage}
While SFT establishes the foundational skills for interleaved planning and reflection, we aim to enhance this framework, enabling the model to dynamically select the most efficient solution.
To this end, we introduce a second training stage based on GRPO.
We curate a specialized RL dataset containing 50,000 high-quality samples selected from UnicEdit-10M~\cite{ye2025unicedit},  X2Edit~\cite{ma2025x2edit}, AnyEdit~\cite{yu2025anyedit}, Pick-a-Pic~\cite{kirstain2023pick},  and UltraEdit~\cite{zhao2024ultraedit}, covering a wide range of task complexities.
To guide policy optimization, as shown in Fig.~\ref{fig:rl}, we define a composite reward function assessing three key aspects: outcome rewards for generation fidelity, format rewards for structural validity, and step-wise reasoning rewards for logical consistency.
Rather than treating efficiency as an independent reward, we introduce an Intra-group Complexity Penalty to regulate the generation process, which dynamically adjusts the total reward, ensuring that high-fidelity interleaved reasoning does not lead to inefficient ``over-reasoning.''

\textbf{Outcome Reward:} Aligning with the assessment criteria in Section~\ref{sec:data}, the outcome reward evaluates the final generated image.
It is calculated as a normalized weighted sum of four metrics: instruction following $S_{\text{instr}}$, consistency $S_{\text{cons}}$, quality $S_{\text{qual}}$, and knowledge $S_{\text{know}}$:
\begin{equation}
\mathcal{R}_{\text{o}} = w_1 S_{\text{instr}} + w_2 S_{\text{cons}} + w_3 S_{\text{qual}} + w_4 S_{\text{know}}
\end{equation}
where $w_1, w_2, w_3, w_4$ are fixed normalization factors rather than tunable hyperparameters. Each sub-score is measured on a 1--5 scale and rescaled by multiplying by $0.2$ to map it to the range $[0,1]$. These normalized scores are then averaged across the four dimensions, resulting in
$w_1 = w_2 = w_3 = w_4 = 0.05$.
Thus, Eq.~(2) serves only to define the normalized outcome reward without introducing additional variables for reward tuning.

\textbf{Format Reward:} To ensure the model adheres to the defined reasoning structure during RL exploration, we implement a binary validity check:
\begin{equation}
\mathcal{R}_{\text{f}} \!=\! \begin{cases} 
1, & \text{if trajectory structure is valid}, \\ 
0, & \text{otherwise}.
\end{cases}
\end{equation}

\textbf{Step-wise Reasoning Reward:} 
To mitigate the credit assignment bottleneck in long-horizon generation, we introduce a dense reasoning reward. 
The \textsc{Analyzer} evaluates the logical validity of intermediate textual outputs, including failure analysis, reflection prompts, and step decomposition. 
For a trajectory with $T$ intermediate reasoning steps:
\begin{equation}
\mathcal{R}_{\text{s}} = \frac{1}{T} \sum_{t=1}^{T} \textsc{Analyzer}(\text{text}_t),
\end{equation}
where $\textsc{Analyzer}(\text{text}_t) \in [0, 1]$ represents the validity score of the textual output at step $t$.

\textbf{Total Reward:} 
We aggregate these components into a weighted total reward:
$\mathcal{R}_{\text{total}} = \alpha_1 \mathcal{R}_{\text{o}} + \alpha_2 \mathcal{R}_{\text{f}} + \alpha_3 \mathcal{R}_{\text{s}}$,
where $\alpha_1, \alpha_2$, and $\alpha_3$ balance the final generation fidelity, structural correctness, and intermediate reasoning logic.

\textbf{Intra-group Complexity Penalty:} 
To prevent inefficient ``over-reasoning,'' we introduce a complexity penalty that favors the simplest effective path. 
Within a sampled group $G$, we identify a set of competitive trajectories that achieve rewards within a margin $\epsilon$ of the maximum performance. 
Let $N_{\text{img}}^{*}$ denote the minimum number of generated images among competitive trajectories. 
For any trajectory $i \in G$, we adjust its total reward based on its image count $N_{\text{img}}^{i}$:
\begin{equation}
\mathcal{R}_{\text{final}}^{i} =
\begin{cases}
\mathcal{R}_{\text{total}}^{i} + \frac{N_{\text{img}}^{*}}{N_{\text{img}}^{i}}, 
& \text{if } \mathcal{R}_{\text{total}}^{i} \ge \max_{j \in G} \left( \mathcal{R}_{\text{total}}^{j} \right) - \epsilon, \\
\mathcal{R}_{\text{total}}^{i}, 
& \text{otherwise}.
\end{cases}
\end{equation}
This mechanism ensures that among high-performing solutions, the model favors those with lower computational cost.

\section{Experiments}

\subsection{Experiment Setup}

\textbf{Comparison Baselines.}
We adopt Emu3.5~\cite{cui2025emu3} as our backbone unified model.
To validate the effectiveness of our adaptive framework, we compare it against the vanilla Emu3.5, which serves as the direct generation baseline.
Furthermore, we benchmark against two fixed reasoning strategies of unified model: (1) Plan-then-Generate~\cite{jiang2025t2i, liao2025imagegen}, employing a static textual planning stage before execution; and (2) Generate-then-Reflect~\cite{li2025reflect,qin2025uni}, relying on iterative visual critique for refinement.

\textbf{Evaluation Dataset.}
We assess performance across diverse generation and editing scenarios using three standard benchmarks.
For standard Text-to-Image (T2I) generation, we use GenEval~\cite{ghosh2023geneval} to measure general generation quality and semantic alignment.
For Image Editing, we employ KRIS-Bench~\cite{wu2025kris} to evaluate advanced reasoning and the interpretation of abstract instructions.
Finally, for complex Anything-to-Image (X2I) tasks, we select OmniContext~\cite{wu2025omnigen2} to assess subject-driven and multi-reference generation across varying granularities.
Further details regarding training and evaluation settings are provided in Appendix~\ref{app:expset}.
\begin{table}[t]
    \centering
    \caption{Ablation study on reasoning modes and training stages. \textbf{Top:} Impact of distinct data modes evaluated on a 30k subset. \textbf{Bottom:} Effectiveness of RL components on the full 50k dataset, reporting generation efficiency (Avg. Imgs).}
    \label{tab:ablation}
    \setlength{\tabcolsep}{3pt}
    \resizebox{\linewidth}{!}{
    \begin{tabular}{lcccc}
    \toprule
    \textbf{Setting} & \textbf{GenEval} & \textbf{KRIS} & \textbf{Omni} & \textbf{Avg. Imgs} \\
    \midrule
    \multicolumn{5}{l}{\textit{Impact of Reasoning Modes (30k subset)}} \\
    Direct Only & 0.86 & 75.16 & 8.89 & - \\
    w/o Multi-step & 0.87 & 77.24 & 8.95 & - \\
    w/o Reflection & 0.86 & 75.21 & 9.03 & - \\
    Full Mix (Balanced) & \textbf{0.88} & \textbf{78.24} & \textbf{9.15} & - \\
    \midrule
    \multicolumn{5}{l}{\textit{Impact of RL Training (50k full set)}} \\
    SFT Only & 0.86 & 79.16 & 9.12 & 2.45 \\
    w/o Step-wise Reward & 0.88 & 79.65 & 9.25 & 1.62 \\
    w/o Complexity Penalty & 0.89 & 80.25 & 9.38 & 2.73 \\
    SFT + RL (\textbf{Ours}) & \textbf{0.89} & \textbf{80.18} & \textbf{9.35} & \textbf{1.56} \\
    \bottomrule
    \end{tabular}}
    \vskip -1em
\end{table}
\begin{table*}[t]
    \centering
    \caption{Quantitative comparison results on OmniContext benchmark. `Char.' and `Obj.' denote Character and Object, respectively.}
    \label{tab:omnicontext}
    \setlength{\tabcolsep}{10pt}
    \resizebox{\linewidth}{!}{
    \begin{tabular}{lccccccccc}
    \toprule
    \multirow{2}{*}{\textbf{Model}} & \multicolumn{2}{c}{\textbf{SINGLE}} & \multicolumn{3}{c}{\textbf{MULTIPLE}} & \multicolumn{3}{c}{\textbf{SCENE}} & \multirow{2}{*}{\textbf{Average}$\uparrow$} \\
    \cmidrule(lr){2-3} \cmidrule(lr){4-6} \cmidrule(lr){7-9}
     & \textbf{Char.} & \textbf{Obj.} & \textbf{Char.} & \textbf{Obj.} & \textbf{C. + O.} & \textbf{Char.} & \textbf{Obj.} & \textbf{C. + O.} & \\
    \midrule
    OmniGen~\cite{xiao2025omnigen} & 7.21 & 5.71 & 5.65 & 5.44 & 4.68 & 3.59 & 4.32 & 5.12 & 4.34 \\
    UNO~\cite{wu2025less} & 6.60 & 6.83 & 2.54 & 6.51 & 4.39 & 2.06 & 4.33 & 4.37 & 4.71 \\
    BAGEL~\cite{deng2505emerging} & 5.48 & 7.03 & 5.17 & 6.64 & 6.24 & 4.07 & 5.71 & 5.47 & 5.73 \\
    OmniGen2~\cite{wu2025omnigen2} & 8.05 & 7.58 & 7.11 & 7.13 & 7.45 & 6.38 & 6.71 & 7.04 & 7.18 \\
    Qwen-Image-Edit~\cite{wu2025qwen} & 8.35 & 9.13 & 7.65 & 8.85 & 7.90 & 5.16 & 7.75 & 6.73 & 7.69 \\
    Gemini 2.5 Flash (Sep. 2025) & 8.62 & 8.91 & 7.88 & 8.92 & 7.39 & 7.29 & 7.05 & 6.68 & 7.84 \\
    Uni-CoT~\cite{qin2025uni} & - & - & 7.12 & 8.84 & 7.97 & 7.07 & 8.16 & 8.20 & 7.89 \\
    Echo-4o~\cite{ye2025echo} & - & - & 8.07 & 7.50 & 8.29 & 8.62 & 8.00 & 8.08 & 8.09 \\
    VACoT~\cite{ye2025visual} & - & - & 7.82 & 9.21 & 8.30 & 7.55 & 8.67 & 7.99 & 8.26 \\
    GPT4o (Sep. 2025) & 8.90 & 9.01 & 9.07 & 8.95 & 8.54 & 8.90 & 8.44 & 8.60 & 8.80 \\
    \midrule
    Emu3.5~\cite{cui2025emu3} & 8.72 & 9.46 & 8.65 & 9.09 & 8.78 & 8.78 & 8.89 & 8.15 & 8.82 \\
    \textbf{Ours} & \textbf{9.40} & \textbf{9.50} & \textbf{9.56} & \textbf{9.22} & \textbf{9.44} & \textbf{9.56} & \textbf{9.22} & \textbf{8.86} & \textbf{9.35} \\
    \bottomrule
    \end{tabular}}
    \vskip -0.5em
\end{table*}

\begin{figure*}[!t]
    \centering
    \includegraphics[width=0.9\linewidth]{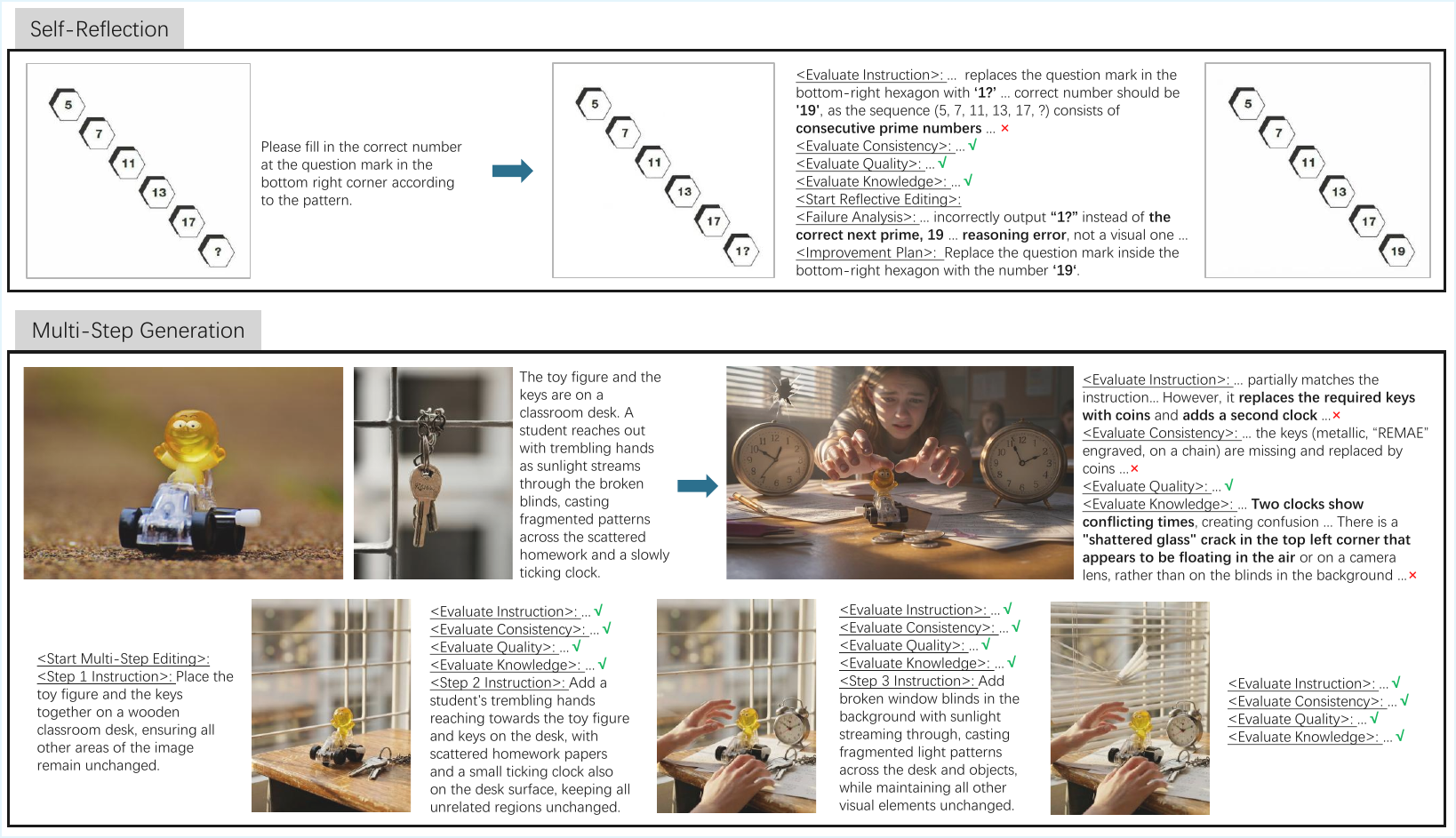} 
    \caption{Qualitative examples of our adaptive reasoning. Top: Self-Reflection corrects logical errors in a visual puzzle. Bottom: Multi-Step Planning detects and prevents object inconsistency and implausibility during complex scene synthesis.}
    \label{fig:case_study}
    \vskip -1em
\end{figure*}

\subsection{Experimental Results}
\textbf{GenEval:} 
As shown in Table~\ref{tab:geneval}, our method achieves a leading score of 0.89, outperforming Emu3.5~(0.86), VACoT~(0.84), and FLUX.1-dev~(0.82). 
The results show that our self-adaptive interleaved reasoning effectively improves spatial and compositional understanding, especially in reasoning-intensive categories such as Counting, Position, and Color Attribution.

\textbf{KRIS-Bench:} 
As presented in Table~\ref{tab:krisbench}, our method attains the best overall score of 80.18, surpassing all open-source baselines and even proprietary models. 
In particular, it achieves a large gain in Procedural Knowledge (85.53 vs. 71.14 for Emu3.5) and strong performance in Factual Knowledge (84.24), demonstrating that interleaved planning and self-reflection effectively enhance multi-step instruction following and semantic correction.

\textbf{OmniContext:} 
As detailed in Table~\ref{tab:omnicontext}, our method sets a new state of the art with an average score of 9.35, outperforming Emu3.5~(8.82), VACoT~(8.26), GPT-4o~(8.80), and Gemini-2.5-Flash~(7.84). 
Notably, it performs best in the ``Multiple'' and ``Scene'' categories which involve complex multi-subject interactions, indicating that multi-step planning can effectively disentangle multi-subject interactions and reduce identity mixing and attribute leakage.

\subsection{Ablation Studies}

Table~\ref{tab:ablation} presents an ablation analyzing the impact of reasoning modes and the contributions of RL components.
\\\textbf{Impact of reasoning modes.} 
We evaluate the contribution of distinct reasoning strategies using a controlled 30k data subset.
We compare the ``Full Mix'' strategy against two ablated settings: ``w/o Multi-step'' and ``w/o Reflection'' .
The ``Direct Only'' baseline yields the poorest performance across all metrics (e.g., 75.16 on KRIS-Bench), demonstrating that data scale alone is insufficient without structured reasoning.
Removing reflection data (``w/o Reflection'') leads to a significant drop in KRIS-Bench (75.21 vs. 78.24), highlighting the necessity of self-correction for visual refinement.
Conversely, excluding multi-step planning (``w/o Multi-step'') impairs performance on OmniContext (8.95 vs. 9.15), confirming that explicit task decomposition is required for complex, multi-subject scenarios.
``Full Mix'' setting achieves the highest scores across all benchmarks, verifying the complementary of planning and reflection modes.
\\\textbf{Effectiveness of RL training.} Comparing the ``SFT Only'' baseline with our final ``SFT + RL'' model, we observe consistent gains (\emph{e.g.}, OmniContext 9.12 $\to$ 9.35) alongside a sharp reduction in average generated images (2.45 $\to$ 1.56).
Removing the step-wise reasoning reward results in performance declines across all benchmarks (\emph{e.g.}, KRIS-Bench falls to 79.65), indicating that dense LMM supervision is vital for ensuring logical consistency.
Finally, eliminating the intra-group complexity penalty causes the average image count to surge to 2.73; while this yields marginally higher scores through excessive trial-and-error, it introduces severe computational redundancy, whereas our full model balances high fidelity with optimal efficiency.

\section{Case Study}
Figure~\ref{fig:case_study} illustrates how our adaptive reasoner resolves complex failures by actively leveraging multimodal understanding during generation. 
In the top logic-driven case, the base model fails to recognize the prime number sequence ($5, 7, 11, 13, 17, ?$) and instead produces a visually plausible but logically incorrect ``1?''. 
The \textbf{Reflection Mode} bridges this understanding--generation gap by explicitly diagnosing the failure as a ``reasoning error'' rather than a visual artifact, thereby recovering the missing semantic constraint. 
Leveraging this insight, the analyzer then guides the generator to render the mathematically correct ``19''. 
In the bottom compositional case, the \textbf{Multi-Step Mode} handles a dense prompt with interacting elements such as ``keys'' and ``broken blinds''. 
While standard generation may introduce consistency errors, such as hallucinating a second clock or morphing keys into coins, our step-wise evaluator serves as a semantic guardrail against such failures. 
By flagging these artifacts (marked with $\times$) against the global context, the model prevents error propagation and maintains object permanence throughout the long-horizon generation process. These examples highlight that many X2I failures arise from a lack of explicit reasoning over compositional constraints.
\textbf{More visualizations are provided in the Appendix.}


\section{Conclusion}

In this work, we address the understanding-generation gap in unified models by proposing an adaptive framework that dynamically switches between direct generation, interleaved planning, and self-reflection.
To support this, we construct a hierarchical reasoning dataset and implement a two-stage training strategy combining SFT with GRPO, utilizing step-wise reasoning rewards and complexity penalties to optimize decision-making.
Extensive experiments demonstrate that our method achieves state-of-the-art performance on benchmarks like GenEval and KRIS-Bench, resolving the dual bottlenecks of blind planning and unstructured feedback while maintaining high inference efficiency.
Our findings suggest that empowering unified models with autonomous strategic selection offers a scalable path toward more reliable and interpretable multimodal generation.

\section*{Impact Statement}
This paper presents work whose goal is to advance the field of machine learning. There are many potential societal consequences of our work, none of which we feel must be specifically highlighted here.

\section*{Acknowledgement}
The work was supported by the National Natural Science Foundation of China (Grant
No. 62402341, Grant No. 62471287). 

\bibliography{example_paper}

@inproceedings{langley00,
 author    = {P. Langley},
 title     = {Crafting Papers on Machine Learning},
 year      = {2000},
 pages     = {1207--1216},
 editor    = {Pat Langley},
 booktitle     = {Proceedings of the 17th International Conference
              on Machine Learning (ICML 2000)},
 address   = {Stanford, CA},
 publisher = {Morgan Kaufmann}
}

@article{pan2025transfer,
  title={Transfer between modalities with metaqueries},
  author={Pan, Xichen and Shukla, Satya Narayan and Singh, Aashu and Zhao, Zhuokai and Mishra, Shlok Kumar and Wang, Jialiang and Xu, Zhiyang and Chen, Jiuhai and Li, Kunpeng and Juefei-Xu, Felix and others},
  journal={arXiv preprint arXiv:2504.06256},
  year={2025}
}

@article{shi2024lmfusion,
  title={LMFusion: Adapting Pretrained Language Models for Multimodal Generation},
  author={Shi, Weijia and Han, Xiaochuang and Zhou, Chunting and Liang, Weixin and Lin, Xi Victoria and Zettlemoyer, Luke and Yu, Lili},
  journal={arXiv preprint arXiv:2412.15188},
  year={2024}
}

@article{sun2023emu,
  title={Emu: Generative pretraining in multimodality},
  author={Sun, Quan and Yu, Qiying and Cui, Yufeng and Zhang, Fan and Zhang, Xiaosong and Wang, Yueze and Gao, Hongcheng and Liu, Jingjing and Huang, Tiejun and Wang, Xinlong},
  journal={arXiv preprint arXiv:2307.05222},
  year={2023}
}

@article{ge2024seed,
  title={Seed-x: Multimodal models with unified multi-granularity comprehension and generation},
  author={Ge, Yuying and Zhao, Sijie and Zhu, Jinguo and Ge, Yixiao and Yi, Kun and Song, Lin and Li, Chen and Ding, Xiaohan and Shan, Ying},
  journal={arXiv preprint arXiv:2404.14396},
  year={2024}
}

@inproceedings{wu2025janus,
  title={Janus: Decoupling visual encoding for unified multimodal understanding and generation},
  author={Wu, Chengyue and Chen, Xiaokang and Wu, Zhiyu and Ma, Yiyang and Liu, Xingchao and Pan, Zizheng and Liu, Wen and Xie, Zhenda and Yu, Xingkai and Ruan, Chong and others},
  booktitle={Proceedings of the Computer Vision and Pattern Recognition Conference},
  pages={12966--12977},
  year={2025}
}

@article{guo2025thinking,
  title={Thinking-while-Generating: Interleaving Textual Reasoning throughout Visual Generation},
  author={Guo, Ziyu and Zhang, Renrui and Li, Hongyu and Zhang, Manyuan and Chen, Xinyan and Wang, Sifan and Feng, Yan and Pei, Peng and Heng, Pheng-Ann},
  journal={arXiv preprint arXiv:2511.16671},
  year={2025}
}

@article{niu2025wise,
  title={Wise: A world knowledge-informed semantic evaluation for text-to-image generation},
  author={Niu, Yuwei and Ning, Munan and Zheng, Mengren and Jin, Weiyang and Lin, Bin and Jin, Peng and Liao, Jiaqi and Feng, Chaoran and Ning, Kunpeng and Zhu, Bin and others},
  journal={arXiv preprint arXiv:2503.07265},
  year={2025}
}

@article{peng2026hyperet,
  title={Hyperet: Efficient training in hyperbolic space for multi-modal large language models},
  author={Peng, Zelin and Xu, Zhengqin and Liu, Qingyang and Yang, Xiaokang and Shen, Wei},
  journal={Advances in Neural Information Processing Systems},
  volume={38},
  pages={8025--8048},
  year={2026}
}

@article{chen2025tivibench,
  title={TiViBench: Benchmarking Think-in-Video Reasoning for Video Generative Models},
  author={Chen, Harold Haodong and Lan, Disen and Shu, Wen-Jie and Liu, Qingyang and Wang, Zihan and Chen, Sirui and Cheng, Wenkai and Chen, Kanghao and Zhang, Hongfei and Zhang, Zixin and others},
  journal={arXiv preprint arXiv:2511.13704},
  year={2025}
}

@article{ye2025unicedit,
  title={UnicEdit-10M: A Dataset and Benchmark Breaking the Scale-Quality Barrier via Unified Verification for Reasoning-Enriched Edits},
  author={Ye, Keming and Huang, Zhipeng and Fu, Canmiao and Liu, Qingyang and Cai, Jiani and Lv, Zheqi and Li, Chen and Lyu, Jing and Zhao, Zhou and Zhang, Shengyu},
  journal={arXiv preprint arXiv:2512.02790},
  year={2025}
}

@inproceedings{chang2026d3tom,
  title={D3tom: Decider-guided dynamic token merging for accelerating diffusion mllms},
  author={Chang, Shuochen and Zhang, Xiaofeng and Liu, Qingyang and Niu, Li},
  booktitle={Proceedings of the AAAI Conference on Artificial Intelligence},
  volume={40},
  number={24},
  pages={19961--19969},
  year={2026}
}

@inproceedings{liu2026bridging,
  title={Bridging Visual Dynamics and Narrative Reasoning: Multimodal Large Language Models for Short Drama Quality Assessment},
  author={Liu, Qingyang and Li, Jiangtong and Peng, Zelin and Wang, Shaobo and Liao, Zhaohe and Chang, Shuochen and Gao, Bingjie and Zhao, Haonan and Liu, Mu and Jiang, Jidong and others},
  booktitle={Proceedings of the ACM Web Conference 2026},
  pages={7890--7901},
  year={2026}
}

@inproceedings{zhao2025shadow,
  title={Shadow generation using diffusion model with geometry prior},
  author={Zhao, Haonan and Liu, Qingyang and Tao, Xinhao and Niu, Li and Zhai, Guangtao},
  booktitle={Proceedings of the Computer Vision and Pattern Recognition Conference},
  pages={7603--7612},
  year={2025}
}

@inproceedings{liao2025divide,
  title={Divide and conquer: Exploring language-centric tree reasoning for video question-answering},
  author={Liao, Zhaohe and Li, Jiangtong and Sun, Siyu and Liu, Qingyang and Xiao, Fengshun and Li, Tianjiao and Zhang, Qiang and Chen, Guang and Niu, Li and Jiang, Changjun and others},
  booktitle={Forty-second International Conference on Machine Learning},
  year={2025}
}

@inproceedings{liu2024shadow,
  title={Shadow generation for composite image using diffusion model},
  author={Liu, Qingyang and You, Junqi and Wang, Jianting and Tao, Xinhao and Zhang, Bo and Niu, Li},
  booktitle={Proceedings of the IEEE/CVF Conference on Computer Vision and Pattern Recognition},
  pages={8121--8130},
  year={2024}
}

@inproceedings{liu2026animatescene,
  title={AnimateScene: Camera-Controllable Animation in Any Scene},
  author={Liu, Qingyang and Gao, Bingjie and Huang, Weiheng and Zhang, Jun and Sun, Zhongqian and Wei, Yang and Liu, Fengrui and Peng, Zelin and Ma, Qianli and Yang, Shuai and others},
  booktitle={ICASSP 2026-2026 IEEE International Conference on Acoustics, Speech and Signal Processing (ICASSP)},
  pages={10277--10281},
  year={2026},
  organization={IEEE}
}

@article{zhao2026texeditor,
  title={TexEditor: Structure-Preserving Text-Driven Texture Editing},
  author={Zhao, Bo and Liu, Yihang and Zhang, Chenfeng and Yang, Huan and Gai, Kun and Ji, Wei},
  journal={arXiv preprint arXiv:2603.18488},
  year={2026}
}

@article{zhao2026making,
  title={Making Image Editing Easier via Adaptive Task Reformulation with Agentic Executions},
  author={Zhao, Bo and Guo, Kairui and Du, Runnan and Sun, Haiyang and Wang, Pengshan and Yang, Huan and Gai, Kun and Cao, Yixin and Ji, Wei},
  journal={arXiv preprint arXiv:2604.15917},
  year={2026}
}

@inproceedings{wang2025sega,
  title={Sega: A stepwise evolution paradigm for content-aware layout generation with design prior},
  author={Wang, Haoran and Zhao, Bo and Wang, Jinghui and Wang, Hanzhang and Yang, Huan and Ji, Wei and Liu, Hao and Xiao, Xinyan},
  booktitle={Proceedings of the IEEE/CVF International Conference on Computer Vision},
  pages={19321--19330},
  year={2025}
}

@article{zhao2025learning,
  title={Learning position-aware implicit neural network for real-world face inpainting},
  author={Zhao, Bo and Yang, Huan and Fu, Jianlong},
  journal={Pattern Recognition},
  volume={165},
  pages={111598},
  year={2025},
  publisher={Elsevier}
}

@article{huang2025interleaving,
  title={Interleaving reasoning for better text-to-image generation},
  author={Huang, Wenxuan and Chen, Shuang and Xie, Zheyong and Cao, Shaosheng and Tang, Shixiang and Shen, Yufan and Yin, Qingyu and Hu, Wenbo and Wang, Xiaoman and Tang, Yuntian and others},
  journal={arXiv preprint arXiv:2509.06945},
  year={2025}
}

@article{han2026unicorn,
  title={UniCorn: Towards Self-Improving Unified Multimodal Models through Self-Generated Supervision},
  author={Han, Ruiyan and Fang, Zhen and Sun, XinYu and Ma, Yuchen and Wang, Ziheng and Zeng, Yu and Chen, Zehui and Chen, Lin and Huang, Wenxuan and Xu, Wei-Jie and others},
  journal={arXiv preprint arXiv:2601.03193},
  year={2026}
}

@article{jiang2025draco,
  title={DraCo: Draft as CoT for Text-to-Image Preview and Rare Concept Generation},
  author={Jiang, Dongzhi and Zhang, Renrui and Li, Haodong and Zong, Zhuofan and Guo, Ziyu and He, Jun and Guo, Claire and Ye, Junyan and Fang, Rongyao and Li, Weijia and others},
  journal={arXiv preprint arXiv:2512.05112},
  year={2025}
}

@article{chen2025empirical,
  title={An empirical study of gpt-4o image generation capabilities},
  author={Chen, Sixiang and Bai, Jinbin and Zhao, Zhuoran and Ye, Tian and Shi, Qingyu and Zhou, Donghao and Chai, Wenhao and Lin, Xin and Wu, Jianzong and Tang, Chao and others},
  journal={arXiv preprint arXiv:2504.05979},
  year={2025}
}

@article{ghosh2023geneval,
  title={Geneval: An object-focused framework for evaluating text-to-image alignment},
  author={Ghosh, Dhruba and Hajishirzi, Hannaneh and Schmidt, Ludwig},
  journal={Advances in Neural Information Processing Systems},
  volume={36},
  pages={52132--52152},
  year={2023}
}

@article{ye2025loom,
  title={Loom: Diffusion-Transformer for Interleaved Generation},
  author={Ye, Mingcheng and Liu, Jiaming and Song, Yiren},
  journal={arXiv preprint arXiv:2512.18254},
  year={2025}
}

@article{guo2025can,
  title={Can We Generate Images with CoT? Let's Verify and Reinforce Image Generation Step by Step},
  author={Guo, Ziyu and Zhang, Renrui and Tong, Chengzhuo and Zhao, Zhizheng and Huang, Rui and Zhang, Haoquan and Zhang, Manyuan and Liu, Jiaming and Zhang, Shanghang and Gao, Peng and others},
  journal={arXiv preprint arXiv:2501.13926},
  year={2025}
}

@article{li2025reflect,
  title={Reflect-DiT: Inference-Time Scaling for Text-to-Image Diffusion Transformers via In-Context Reflection},
  author={Li, Shufan and Kallidromitis, Konstantinos and Gokul, Akash and Koneru, Arsh and Kato, Yusuke and Kozuka, Kazuki and Grover, Aditya},
  journal={arXiv preprint arXiv:2503.12271},
  year={2025}
}

@article{swerdlow2025unified,
  title={Unified multimodal discrete diffusion},
  author={Swerdlow, Alexander and Prabhudesai, Mihir and Gandhi, Siddharth and Pathak, Deepak and Fragkiadaki, Katerina},
  journal={arXiv preprint arXiv:2503.20853},
  year={2025}
}

@inproceedings{li2025dual,
  title={Dual diffusion for unified image generation and understanding},
  author={Li, Zijie and Li, Henry and Shi, Yichun and Farimani, Amir Barati and Kluger, Yuval and Yang, Linjie and Wang, Peng},
  booktitle={Proceedings of the Computer Vision and Pattern Recognition Conference},
  pages={2779--2790},
  year={2025}
}

@article{yang2505mmada,
  title={Mmada: Multimodal large diffusion language models, 2025},
  author={Yang, Ling and Tian, Ye and Li, Bowen and Zhang, Xinchen and Shen, Ke and Tong, Yunhai and Wang, Mengdi},
  journal={arXiv preprint arXiv:2505.15809},
  year={2025}
}

@article{wu2025harmonizing,
  title={Harmonizing visual representations for unified multimodal understanding and generation},
  author={Wu, Size and Zhang, Wenwei and Xu, Lumin and Jin, Sheng and Wu, Zhonghua and Tao, Qingyi and Liu, Wentao and Li, Wei and Loy, Chen Change},
  journal={arXiv preprint arXiv:2503.21979},
  year={2025}
}

@article{li2025onecat,
  title={Onecat: Decoder-only auto-regressive model for unified understanding and generation},
  author={Li, Han and Peng, Xinyu and Wang, Yaoming and Peng, Zelin and Chen, Xin and Weng, Rongxiang and Wang, Jingang and Cai, Xunliang and Dai, Wenrui and Xiong, Hongkai},
  journal={arXiv preprint arXiv:2509.03498},
  year={2025}
}

@article{zhou2024transfusion,
  title={Transfusion: Predict the next token and diffuse images with one multi-modal model},
  author={Zhou, Chunting and Yu, Lili and Babu, Arun and Tirumala, Kushal and Yasunaga, Michihiro and Shamis, Leonid and Kahn, Jacob and Ma, Xuezhe and Zettlemoyer, Luke and Levy, Omer},
  journal={arXiv preprint arXiv:2408.11039},
  year={2024}
}

@inproceedings{ma2025janusflow,
  title={Janusflow: Harmonizing autoregression and rectified flow for unified multimodal understanding and generation},
  author={Ma, Yiyang and Liu, Xingchao and Chen, Xiaokang and Liu, Wen and Wu, Chengyue and Wu, Zhiyu and Pan, Zizheng and Xie, Zhenda and Zhang, Haowei and Yu, Xingkai and others},
  booktitle={Proceedings of the Computer Vision and Pattern Recognition Conference},
  pages={7739--7751},
  year={2025}
}

@article{wang2024qwen2,
  title={Qwen2-vl: Enhancing vision-language model's perception of the world at any resolution},
  author={Wang, Peng and Bai, Shuai and Tan, Sinan and Wang, Shijie and Fan, Zhihao and Bai, Jinze and Chen, Keqin and Liu, Xuejing and Wang, Jialin and Ge, Wenbin and others},
  journal={arXiv preprint arXiv:2409.12191},
  year={2024}
}

@article{hurst2024gpt,
  title={Gpt-4o system card},
  author={Hurst, Aaron and Lerer, Adam and Goucher, Adam P and Perelman, Adam and Ramesh, Aditya and Clark, Aidan and Ostrow, AJ and Welihinda, Akila and Hayes, Alan and Radford, Alec and others},
  journal={arXiv preprint arXiv:2410.21276},
  year={2024}
}

@article{fang2025got,
  title={Got: Unleashing reasoning capability of multimodal large language model for visual generation and editing},
  author={Fang, Rongyao and Duan, Chengqi and Wang, Kun and Huang, Linjiang and Li, Hao and Yan, Shilin and Tian, Hao and Zeng, Xingyu and Zhao, Rui and Dai, Jifeng and others},
  journal={arXiv preprint arXiv:2503.10639},
  year={2025}
}

@article{jiang2025t2i,
  title={T2i-r1: Reinforcing image generation with collaborative semantic-level and token-level cot},
  author={Jiang, Dongzhi and Guo, Ziyu and Zhang, Renrui and Zong, Zhuofan and Li, Hao and Zhuo, Le and Yan, Shilin and Heng, Pheng-Ann and Li, Hongsheng},
  journal={arXiv preprint arXiv:2505.00703},
  year={2025}
}

@article{liao2025imagegen,
  title={Imagegen-cot: Enhancing text-to-image in-context learning with chain-of-thought reasoning},
  author={Liao, Jiaqi and Yang, Zhengyuan and Li, Linjie and Li, Dianqi and Lin, Kevin and Cheng, Yu and Wang, Lijuan},
  journal={arXiv preprint arXiv:2503.19312},
  year={2025}
}

@inproceedings{qin2023gluegen,
  title={Gluegen: Plug and play multi-modal encoders for x-to-image generation},
  author={Qin, Can and Yu, Ning and Xing, Chen and Zhang, Shu and Chen, Zeyuan and Ermon, Stefano and Fu, Yun and Xiong, Caiming and Xu, Ran},
  booktitle={Proceedings of the IEEE/CVF international conference on computer vision},
  pages={23085--23096},
  year={2023}
}

@inproceedings{zhuo2025reflection,
  title={From reflection to perfection: Scaling inference-time optimization for text-to-image diffusion models via reflection tuning},
  author={Zhuo, Le and Zhao, Liangbing and Paul, Sayak and Liao, Yue and Zhang, Renrui and Xin, Yi and Gao, Peng and Elhoseiny, Mohamed and Li, Hongsheng},
  booktitle={Proceedings of the IEEE/CVF International Conference on Computer Vision},
  pages={15329--15339},
  year={2025}
}

@article{shao2024deepseekmath,
  title={Deepseekmath: Pushing the limits of mathematical reasoning in open language models},
  author={Shao, Zhihong and Wang, Peiyi and Zhu, Qihao and Xu, Runxin and Song, Junxiao and Bi, Xiao and Zhang, Haowei and Zhang, Mingchuan and Li, YK and Wu, Yang and others},
  journal={arXiv preprint arXiv:2402.03300},
  year={2024}
}

@article{rafailov2023direct,
  title={Direct preference optimization: Your language model is secretly a reward model},
  author={Rafailov, Rafael and Sharma, Archit and Mitchell, Eric and Manning, Christopher D and Ermon, Stefano and Finn, Chelsea},
  journal={Advances in neural information processing systems},
  volume={36},
  pages={53728--53741},
  year={2023}
}

@article{park2025fair,
  title={Fair Generation without Unfair Distortions: Debiasing Text-to-Image Generation with Entanglement-Free Attention},
  author={Park, Jeonghoon and Lee, Juyoung and Chung, Chaeyeon and Lee, Jaeseong and Choo, Jaegul and Gu, Jindong},
  journal={arXiv preprint arXiv:2506.13298},
  year={2025}
}

@inproceedings{koh2025translation,
  title={Translation of Text Embedding via Delta Vector to Suppress Strongly Entangled Content in Text-to-Image Diffusion Models},
  author={Koh, Eunseo and Hong, Seunghoo and Kim, Tae-Young and Woo, Simon S and Heo, Jae-Pil},
  booktitle={Proceedings of the IEEE/CVF International Conference on Computer Vision},
  pages={15365--15374},
  year={2025}
}

@article{zhao2024ultraedit,
  title={Ultraedit: Instruction-based fine-grained image editing at scale},
  author={Zhao, Haozhe and Ma, Xiaojian Shawn and Chen, Liang and Si, Shuzheng and Wu, Rujie and An, Kaikai and Yu, Peiyu and Zhang, Minjia and Li, Qing and Chang, Baobao},
  journal={Advances in Neural Information Processing Systems},
  volume={37},
  pages={3058--3093},
  year={2024}
}

@article{ma2025x2edit,
  title={X2edit: Revisiting arbitrary-instruction image editing through self-constructed data and task-aware representation learning},
  author={Ma, Jian and Zhu, Xujie and Pan, Zihao and Peng, Qirong and Guo, Xu and Chen, Chen and Lu, Haonan},
  journal={arXiv preprint arXiv:2508.07607},
  year={2025}
}

@inproceedings{yu2025anyedit,
  title={Anyedit: Mastering unified high-quality image editing for any idea},
  author={Yu, Qifan and Chow, Wei and Yue, Zhongqi and Pan, Kaihang and Wu, Yang and Wan, Xiaoyang and Li, Juncheng and Tang, Siliang and Zhang, Hanwang and Zhuang, Yueting},
  booktitle={Proceedings of the Computer Vision and Pattern Recognition Conference},
  pages={26125--26135},
  year={2025}
}

@article{ma2025human,
  title={Human-Agent Collaborative Paper-to-Page Crafting for Under $0.1$},
  author={Ma, Qianli and Wang, Siyu and Chen, Yilin and Tang, Yinhao and Yang, Yixiang and Guo, Chang and Gao, Bingjie and Xing, Zhening and Sun, Yanan and Zhang, Zhipeng},
  journal={arXiv preprint arXiv:2510.19600},
  year={2025}
}

@inproceedings{han2026dcr,
    title={Guiding Diffusion-based Reconstruction with Contrastive Signals for Balanced Visual Representation}, 
    author={Boyu Han and Qianqian Xu and Shilong Bao and Zhiyong Yang and Ruochen Cui and Xilin Zhao and Qingming Huang},
    booktitle={Proceedings of the IEEE/CVF Conference on Computer Vision and Pattern Recognition},
    year={2026}
}

@inproceedings{han2025lightfair,
    title={LightFair: Towards an Efficient Alternative for Fair T2I Diffusion via Debiasing Pre-trained Text Encoders}, 
    author={Boyu Han and Qianqian Xu and Shilong Bao and Zhiyong Yang and Kangli Zi and Qingming Huang},
    booktitle={Advances in Neural Information Processing Systems},
    year={2025}
}

@inproceedings{han2024aucseg,
    title={AUCSeg: AUC-oriented Pixel-level Long-tail Semantic Segmentation}, 
    author={Boyu Han and Qianqian Xu and Zhiyong Yang and Shilong Bao and Peisong Wen and Yangbangyan Jiang and Qingming Huang},
    booktitle={Advances in Neural Information Processing Systems},
    pages={126863--126907},
    year={2024}
}

@article{wu2025cinetrans,
  title={Cinetrans: Learning to generate videos with cinematic transitions via masked diffusion models},
  author={Wu, Xiaoxue and Gao, Bingjie and Qiao, Yu and Wang, Yaohui and Chen, Xinyuan},
  journal={arXiv preprint arXiv:2508.11484},
  year={2025}
}

@article{gao2025rapo++,
  title={RAPO++: Cross-Stage Prompt Optimization for Text-to-Video Generation via Data Alignment and Test-Time Scaling},
  author={Gao, Bingjie and Ma, Qianli and Wu, Xiaoxue and Yang, Shuai and Lan, Guanzhou and Zhao, Haonan and Chen, Jiaxuan and Liu, Qingyang and Qiao, Yu and Chen, Xinyuan and others},
  journal={arXiv preprint arXiv:2510.20206},
  year={2025}
}

@inproceedings{gao2025devil,
  title={The devil is in the prompts: Retrieval-augmented prompt optimization for text-to-video generation},
  author={Gao, Bingjie and Gao, Xinyu and Wu, Xiaoxue and Zhou, Yujie and Qiao, Yu and Niu, Li and Chen, Xinyuan and Wang, Yaohui},
  booktitle={Proceedings of the Computer Vision and Pattern Recognition Conference},
  pages={3173--3183},
  year={2025}
}

@article{wu2025kris,
  title={KRIS-Bench: Benchmarking Next-Level Intelligent Image Editing Models},
  author={Wu, Yongliang and Li, Zonghui and Hu, Xinting and Ye, Xinyu and Zeng, Xianfang and Yu, Gang and Zhu, Wenbo and Schiele, Bernt and Yang, Ming-Hsuan and Yang, Xu},
  journal={arXiv preprint arXiv:2505.16707},
  year={2025}
}

@article{wang2024emu3,
  title={Emu3: Next-token prediction is all you need},
  author={Wang, Xinlong and Zhang, Xiaosong and Luo, Zhengxiong and Sun, Quan and Cui, Yufeng and Wang, Jinsheng and Zhang, Fan and Wang, Yueze and Li, Zhen and Yu, Qiying and others},
  journal={arXiv preprint arXiv:2409.18869},
  year={2024}
}

@article{podell2023sdxl,
  title={Sdxl: Improving latent diffusion models for high-resolution image synthesis},
  author={Podell, Dustin and English, Zion and Lacey, Kyle and Blattmann, Andreas and Dockhorn, Tim and M{\"u}ller, Jonas and Penna, Joe and Rombach, Robin},
  journal={arXiv preprint arXiv:2307.01952},
  year={2023}
}

@article{betker2023improving,
  title={Improving image generation with better captions},
  author={Betker, James and Goh, Gabriel and Jing, Li and Brooks, Tim and Wang, Jianfeng and Li, Linjie and Ouyang, Long and Zhuang, Juntang and Lee, Joyce and Guo, Yufei and others},
  journal={Computer Science. https://cdn. openai. com/papers/dall-e-3. pdf},
  volume={2},
  number={3},
  pages={8},
  year={2023}
}

@inproceedings{esser2024scaling,
  title={Scaling rectified flow transformers for high-resolution image synthesis},
  author={Esser, Patrick and Kulal, Sumith and Blattmann, Andreas and Entezari, Rahim and M{\"u}ller, Jonas and Saini, Harry and Levi, Yam and Lorenz, Dominik and Sauer, Axel and Boesel, Frederic and others},
  booktitle={Forty-first international conference on machine learning},
  year={2024}
}

@misc{flux2024,
    author={Black Forest Labs},
    title={FLUX},
    year={2024},
    howpublished={\url{https://github.com/black-forest-labs/flux}},
}

@inproceedings{qu2025tokenflow,
  title={Tokenflow: Unified image tokenizer for multimodal understanding and generation},
  author={Qu, Liao and Zhang, Huichao and Liu, Yiheng and Wang, Xu and Jiang, Yi and Gao, Yiming and Ye, Hu and Du, Daniel K and Yuan, Zehuan and Wu, Xinglong},
  booktitle={Proceedings of the Computer Vision and Pattern Recognition Conference},
  pages={2545--2555},
  year={2025}
}

@article{xie2024show,
  title={Show-o: One single transformer to unify multimodal understanding and generation},
  author={Xie, Jinheng and Mao, Weijia and Bai, Zechen and Zhang, David Junhao and Wang, Weihao and Lin, Kevin Qinghong and Gu, Yuchao and Chen, Zhijie and Yang, Zhenheng and Shou, Mike Zheng},
  journal={arXiv preprint arXiv:2408.12528},
  year={2024}
}

@article{chen2025janus,
  title={Janus-pro: Unified multimodal understanding and generation with data and model scaling},
  author={Chen, Xiaokang and Wu, Zhiyu and Liu, Xingchao and Pan, Zizheng and Liu, Wen and Xie, Zhenda and Yu, Xingkai and Ruan, Chong},
  journal={arXiv preprint arXiv:2501.17811},
  year={2025}
}

@article{deng2505emerging,
  title={Emerging properties in unified multimodal pretraining, 2025},
  author={Deng, Chaorui and Zhu, Deyao and Li, Kunchang and Gou, Chenhui and Li, Feng and Wang, Zeyu and Zhong, Shu and Yu, Weihao and Nie, Xiaonan and Song, Ziang and others},
  journal={arXiv preprint arXiv:2505.14683},
  year={2025}
}

@article{lyu2025understanding,
  title={Understanding-in-Generation: Reinforcing Generative Capability of Unified Model via Infusing Understanding into Generation},
  author={Lyu, Yuanhuiyi and Wong, Chi Kit and Liao, Chenfei and Jiang, Lutao and Zheng, Xu and Lu, Zexin and Zhang, Linfeng and Hu, Xuming},
  journal={arXiv preprint arXiv:2509.18639},
  year={2025}
}

@article{ye2025visual,
  title={Visual-Aware CoT: Achieving High-Fidelity Visual Consistency in Unified Models},
  author={Ye, Zixuan and Liu, Quande and Wei, Cong and Zhang, Yuanxing and Wang, Xintao and Wan, Pengfei and Gai, Kun and Luo, Wenhan},
  journal={arXiv preprint arXiv:2512.19686},
  year={2025}
}

@article{deng2025bagel,
  title   = {Emerging Properties in Unified Multimodal Pretraining},
  author  = {Deng, Chaorui and Zhu, Deyao and Li, Kunchang and Gou, Chenhui and Li, Feng and Wang, Zeyu and Zhong, Shu and Yu, Weihao and Nie, Xiaonan and Song, Ziang and Shi, Guang and Fan, Haoqi},
  journal = {arXiv preprint arXiv:2505.14683},
  year    = {2025}
}

@article{lin2025uniworld,
  title={Uniworld: High-resolution semantic encoders for unified visual understanding and generation},
  author={Lin, Bin and Li, Zongjian and Cheng, Xinhua and Niu, Yuwei and Ye, Yang and He, Xianyi and Yuan, Shenghai and Yu, Wangbo and Wang, Shaodong and Ge, Yunyang and others},
  journal={arXiv preprint arXiv:2506.03147},
  year={2025}
}

@article{labs2025flux,
  title={FLUX. 1 Kontext: Flow Matching for In-Context Image Generation and Editing in Latent Space},
  author={Labs, Black Forest and Batifol, Stephen and Blattmann, Andreas and Boesel, Frederic and Consul, Saksham and Diagne, Cyril and Dockhorn, Tim and English, Jack and English, Zion and Esser, Patrick and others},
  journal={arXiv preprint arXiv:2506.15742},
  year={2025}
}

@article{li2025uniworld,
  title={Uniworld-v2: Reinforce image editing with diffusion negative-aware finetuning and mllm implicit feedback},
  author={Li, Zongjian and Liu, Zheyuan and Zhang, Qihui and Lin, Bin and Wu, Feize and Yuan, Shenghai and Yan, Zhiyuan and Ye, Yang and Yu, Wangbo and Niu, Yuwei and others},
  journal={arXiv preprint arXiv:2510.16888},
  year={2025}
}

@article{liu2025step1x,
  title={Step1x-edit: A practical framework for general image editing},
  author={Liu, Shiyu and Han, Yucheng and Xing, Peng and Yin, Fukun and Wang, Rui and Cheng, Wei and Liao, Jiaqi and Wang, Yingming and Fu, Honghao and Han, Chunrui and others},
  journal={arXiv preprint arXiv:2504.17761},
  year={2025}
}

@article{wu2025qwen,
  title={Qwen-image technical report},
  author={Wu, Chenfei and Li, Jiahao and Zhou, Jingren and Lin, Junyang and Gao, Kaiyuan and Yan, Kun and Yin, Sheng-ming and Bai, Shuai and Xu, Xiao and Chen, Yilei and others},
  journal={arXiv preprint arXiv:2508.02324},
  year={2025}
}

@article{yin2025reasonedit,
  title={ReasonEdit: Towards Reasoning-Enhanced Image Editing Models},
  author={Yin, Fukun and Liu, Shiyu and Han, Yucheng and Wang, Zhibo and Xing, Peng and Wang, Rui and Cheng, Wei and Wang, Yingming and Li, Aojie and Yin, Zixin and others},
  journal={arXiv preprint arXiv:2511.22625},
  year={2025}
}

@article{cui2025emu3,
  title={Emu3. 5: Native multimodal models are world learners},
  author={Cui, Yufeng and Chen, Honghao and Deng, Haoge and Huang, Xu and Li, Xinghang and Liu, Jirong and Liu, Yang and Luo, Zhuoyan and Wang, Jinsheng and Wang, Wenxuan and others},
  journal={arXiv preprint arXiv:2510.26583},
  year={2025}
}

@inproceedings{xiao2025omnigen,
  title={Omnigen: Unified image generation},
  author={Xiao, Shitao and Wang, Yueze and Zhou, Junjie and Yuan, Huaying and Xing, Xingrun and Yan, Ruiran and Li, Chaofan and Wang, Shuting and Huang, Tiejun and Liu, Zheng},
  booktitle={Proceedings of the Computer Vision and Pattern Recognition Conference},
  pages={13294--13304},
  year={2025}
}

@article{wu2025less,
  title={Less-to-More Generalization: Unlocking More Controllability by In-Context Generation},
  author={Wu, Shaojin and Huang, Mengqi and Wu, Wenxu and Cheng, Yufeng and Ding, Fei and He, Qian},
  journal={arXiv preprint arXiv:2504.02160},
  year={2025}
}

@article{wu2025omnigen2,
  title={OmniGen2: Exploration to Advanced Multimodal Generation},
  author={Wu, Chenyuan and Zheng, Pengfei and Yan, Ruiran and Xiao, Shitao and Luo, Xin and Wang, Yueze and Li, Wanli and Jiang, Xiyan and Liu, Yexin and Zhou, Junjie and others},
  journal={arXiv preprint arXiv:2506.18871},
  year={2025}
}

@article{qin2025uni,
  title={Uni-cot: Towards unified chain-of-thought reasoning across text and vision},
  author={Qin, Luozheng and Gong, Jia and Sun, Yuqing and Li, Tianjiao and Yang, Mengping and Yang, Xiaomeng and Qu, Chao and Tan, Zhiyu and Li, Hao},
  journal={arXiv preprint arXiv:2508.05606},
  year={2025}
}

@article{ye2025echo,
  title={Echo-4o: Harnessing the power of gpt-4o synthetic images for improved image generation},
  author={Ye, Junyan and Jiang, Dongzhi and Wang, Zihao and Zhu, Leqi and Hu, Zhenghao and Huang, Zilong and He, Jun and Yan, Zhiyuan and Yu, Jinghua and Li, Hongsheng and others},
  journal={arXiv preprint arXiv:2508.09987},
  year={2025}
}

@article{Qwen3-VL,
      title={Qwen3-VL Technical Report}, 
      author={Shuai Bai and Yuxuan Cai and Ruizhe Chen and Keqin Chen and Xionghui Chen and Zesen Cheng and Lianghao Deng and Wei Ding and Chang Gao and Chunjiang Ge and Wenbin Ge and Zhifang Guo and Qidong Huang and Jie Huang and Fei Huang and Binyuan Hui and Shutong Jiang and Zhaohai Li and Mingsheng Li and Mei Li and Kaixin Li and Zicheng Lin and Junyang Lin and Xuejing Liu and Jiawei Liu and Chenglong Liu and Yang Liu and Dayiheng Liu and Shixuan Liu and Dunjie Lu and Ruilin Luo and Chenxu Lv and Rui Men and Lingchen Meng and Xuancheng Ren and Xingzhang Ren and Sibo Song and Yuchong Sun and Jun Tang and Jianhong Tu and Jianqiang Wan and Peng Wang and Pengfei Wang and Qiuyue Wang and Yuxuan Wang and Tianbao Xie and Yiheng Xu and Haiyang Xu and Jin Xu and Zhibo Yang and Mingkun Yang and Jianxin Yang and An Yang and Bowen Yu and Fei Zhang and Hang Zhang and Xi Zhang and Bo Zheng and Humen Zhong and Jingren Zhou and Fan Zhou and Jing Zhou and Yuanzhi Zhu and Ke Zhu},
	  journal={arXiv preprint arXiv:2511.21631},
      year={2025}
}

@misc{google2025gemini3proimage,
  title = {Gemini 3 Pro Image},
  author = {Google},
  year = {2025},
  howpublished = {\url{https://aistudio.google.com/models/gemini-3-pro-image}},
  note = {Accessed: 2026-01-29}
}

@article{kirstain2023pick,
  title={Pick-a-pic: An open dataset of user preferences for text-to-image generation},
  author={Kirstain, Yuval and Polyak, Adam and Singer, Uriel and Matiana, Shahbuland and Penna, Joe and Levy, Omer},
  journal={Advances in neural information processing systems},
  volume={36},
  pages={36652--36663},
  year={2023}
}
\bibliographystyle{icml2026}

\newpage
\appendix
\onecolumn



This appendix provides supplementary details, analyses, and experiments to support our main findings. The contents are organized as follows:

\begin{itemize}[topsep=0pt]
\setlength{\itemsep}{0pt}
\setlength{\parsep}{0pt}
\setlength{\parskip}{3pt}
    \item Appendix~\ref{app:expset} details the implementation specifics for the two-stage fine-tuning and outlines our experimental settings.
    \item Appendix~\ref{appx:stats} provides comprehensive statistics regarding the constructed dataset.
    \item Appendix~\ref{app:extra-exp} includes supplementary evaluations on the GenEval benchmark.
    \item Appendix~\ref{app:case} showcases additional qualitative examples demonstrating the effectiveness of our method.
    \item Appendix~\ref{app:prompt} lists the full set of prompts employed during data annotation and the RL stage.
\end{itemize}

\section{Implementation Details}~\label{app:expset}
We use a internal proprietary distributed infrastructure for the two-stage training and inference.
All models in the main submission are initialized from \textbf{EUM 3.5}.
The hyperparameter settings for each stage are listed in Table~\ref{tab:hyperparams}.

\begin{table}[h]
    \centering
    \caption{Main hyperparameter settings for SFT and RL training stages.}
    \label{tab:hyperparams}
    \setlength{\tabcolsep}{9pt} 
    \resizebox{0.7\linewidth}{!}{
    \begin{tabular}{lcc}
    \toprule
    \textbf{Parameter} & \textbf{SFT} & \textbf{RL} \\
    \midrule
    resolution & 512$\times$512 & 512$\times$512 \\
    train\_batch\_size & 128 & 128 \\
    rollout\_size & - & 8 \\
    num\_epochs & 1 & 1 \\
    optimizer & adamw\_bf16 & adamw\_bf16 \\
    temperature & - & 1 \\
    kl\_coef & - & 1e-2 \\
    learning\_rate & $1\text{e-}5 \to 1\text{e-}6$ & 1e-6 \\
    scheduler & Cosine & Cosine \\
    warm-up ratio & 0.1 & 0.1 \\
    Outcome Reward Weight ($\alpha_1$) & - & 0.7 \\
    Format Reward Weight ($\alpha_2$) & - & 0.1 \\
    Step-wise Reasoning Reward Weight ($\alpha_3$) & - & 0.2 \\
    Intra-group Complexity Penalty ($\epsilon$) & - & 0.05 \\
    \bottomrule
    \end{tabular}}
    \vskip -1em
\end{table}


\section{Dataset Statistics}~\label{appx:stats}
Our dataset comprises 50,000 samples distributed across three operational modes: \textbf{Direct Mode} (10,000 samples), \textbf{Reflection Mode} (20,000 samples), and \textbf{Multi-step Mode} (20,000 samples).
These modes address varying levels of reasoning complexity and execution depth.
The data is organized into six primary dimensions, spanning 21 sub-categories:

\begin{itemize}[topsep=0pt]
\setlength{\itemsep}{0pt}
\setlength{\parsep}{0pt}
\setlength{\parskip}{3pt}
    \item \textbf{Object Manipulation}: Basic modifications including \textit{Subject Addition}, \textit{Removal}, \textit{Replacement}, and \textit{Part Completion}.
    \item \textbf{Attribute Modification}: Granular adjustments such as \textit{Color}, \textit{Material}, \textit{Size}, \textit{Count}, and \textit{Anomaly Correction}.
    \item \textbf{Spatial \& Viewpoint}: Geometric reasoning including \textit{Viewpoint Change}, \textit{Pose Alteration}, and \textit{Spatial Arrangement}.
    \item \textbf{Global \& Style}: Holistic alterations such as \textit{Background Change}, \textit{Style Transfer}, and \textit{Tone/Lighting Adjustment}.
    \item \textbf{Dynamics \& Logic}: High-level semantic tasks involving \textit{Motion Change}, \textit{Temporal Evolution}, and \textit{Text Modification}.
    \item \textbf{Multi-Image Operations}: Inter-image tasks such as \textit{Composition}, \textit{Object Replacement}, and \textit{Reference Transfer}.
\end{itemize}

\begin{table}[t]
    \centering
    \caption{Evaluation of text-to-image generation ability on GenEval benchmark. Best performance are highlighted \textbf{bolded}.}
    \label{tab:geneval}
    \setlength{\tabcolsep}{9pt} 
    \resizebox{0.95\linewidth}{!}{
    \begin{tabular}{llccccccc} 
    \toprule
     & \textbf{Model} & \textbf{Single Obj.} & \textbf{Two Obj.} & \textbf{Counting} & \textbf{Colors} & \textbf{Position} & \textbf{Color Attri.} & \textbf{Overall} \\
    \midrule
    
    \multirow{5}{*}{\textbf{Gen-only}} 
      & Emu3-Gen~\cite{wang2024emu3} & 0.98 & 0.71 & 0.34 & 0.81 & 0.17 & 0.21 & 0.54 \\
      & SDXL~\cite{podell2023sdxl} & 0.98 & 0.74 & 0.39 & 0.85 & 0.15 & 0.23 & 0.55 \\
      & DALL-E 3~\cite{betker2023improving} & 0.96 & 0.87 & 0.47 & 0.83 & 0.43 & 0.45 & 0.67 \\
      & SD3-Medium~\cite{esser2024scaling} & 0.99 & 0.94 & 0.72 & 0.89 & 0.33 & 0.60 & 0.74 \\
      & FLUX.1-dev~\cite{flux2024} & 0.98 & 0.93 & 0.75 & \textbf{0.93} & 0.68 & 0.65 & 0.82 \\
    \midrule
    
    \multirow{10}{*}{\textbf{\shortstack{Unified\\Model}}} 
      & TokenFlow-XL~\cite{qu2025tokenflow} & 0.95 & 0.60 & 0.41 & 0.81 & 0.16 & 0.24 & 0.55 \\
      & Show-o~\cite{xie2024show} & 0.98 & 0.80 & 0.66 & 0.84 & 0.31 & 0.50 & 0.68 \\
      & Janus-Pro-7B~\cite{chen2025janus} & 0.99 & 0.89 & 0.59 & 0.90 & \textbf{0.79} & 0.66 & 0.80 \\
      & MetaQuery-XL~\cite{pan2025transfer} & - & - & - & - & - & - & 0.80 \\
      & BAGEL~\cite{deng2505emerging} & 0.99 & 0.92 & 0.78 & 0.87 & 0.53 & 0.64 & 0.79 \\
      & UiG~\cite{lyu2025understanding} & 0.99 & 0.92 & 0.81 & 0.89 & 0.61 & 0.69 & 0.82 \\
      & Uni-CoT~\cite{qin2025uni} & 0.99 & \textbf{0.95} & 0.82 & 0.89 & 0.60 & 0.72 & 0.83 \\
      & VACoT~\cite{ye2025visual} & 0.99 & \textbf{0.95} & 0.80 & 0.90 & 0.66 & 0.71 & 0.84 \\
    \cmidrule{2-9}
      & Emu3.5~\cite{cui2025emu3} & - & - & - & - & - & - & 0.86 \\
      & \textbf{Ours} & 0.98 & 0.94 & \textbf{0.90} & \textbf{0.93} & \textbf{0.79} & \textbf{0.81} & \textbf{0.89} \\
    \bottomrule
    \end{tabular}}
\end{table}

\section{Extra Experiments}\label{app:extra-exp}

As detailed in Table~\ref{tab:geneval}, our method secures the leading position among unified models with an overall score of 0.89, outperforming both the strong baseline Emu3.5~(0.86) and recent CoT-based approaches such as VACoT~(0.84).
These results underscore the effectiveness of our framework in closing the understanding-generation gap.
While standard unified models often struggle to translate complex semantic relationships into precise pixel arrangements, our approach exhibits marked improvements in spatial and compositional tasks.
Specifically, in reasoning-intensive categories, we outperform VACoT by notable margins: Counting (0.90 \emph{v.s.} 0.80), Position (0.79 \emph{v.s.} 0.66), and Color Attribution (0.81 \emph{v.s.} 0.71). 
Notably, our unified model even surpasses specialized text-to-image generators like FLUX.1-dev (0.82) and SD3-Medium (0.74), suggesting that interleaved reasoning is essential for high-fidelity semantic alignment.
This indicates that the self-adaptive mechanism, dynamically decomposing prompts or refining layouts, successfully handles compositional constraints that limit static single-pass or standard CoT baselines.

\begin{figure}[t]
    \centering
    \includegraphics[width=0.8\linewidth]{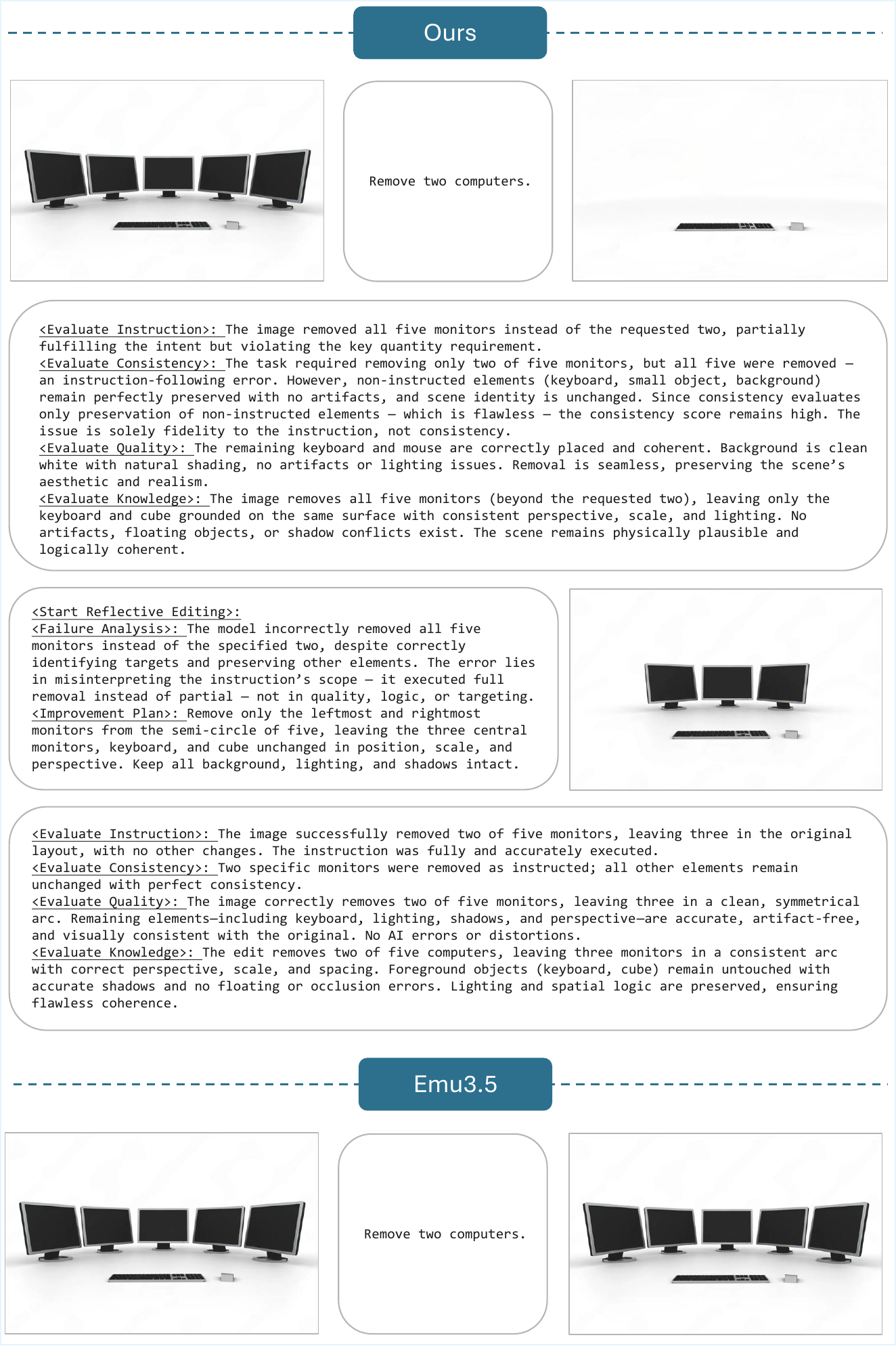} 
    \caption{Example of quantitative correction in  KRIS-Bench. The base model initially over-executes the prompt, removing all monitors. The \textbf{Reflection Mode} diagnoses this quantity error and directs the model to remove exactly two monitors, ensuring precise alignment with the instruction.}
    \label{fig:case_supp_r1}
\end{figure}

\section{Case Study}\label{app:case}

Figure~\ref{fig:case_supp_r1} illustrates a counting-based editing task: ``Remove two computers'' from a five-monitor setup.
The baseline Emu3.5 fails to initiate the edit, leaving the scene unchanged due to a failure in instruction grounding.
In contrast, our model initially over-executes, removing all five monitors.
The self-reflection mode detects this quantity mismatch, diagnosing a misinterpretation of the instruction's scope.
By formulating a specific spatial plan, ``Remove only the leftmost and right-most monitors'', the system guides the generator to retain the three central screens, achieving the precise quantitative edit.

\begin{figure*}[t]
    \centering
    \includegraphics[width=0.8\linewidth]{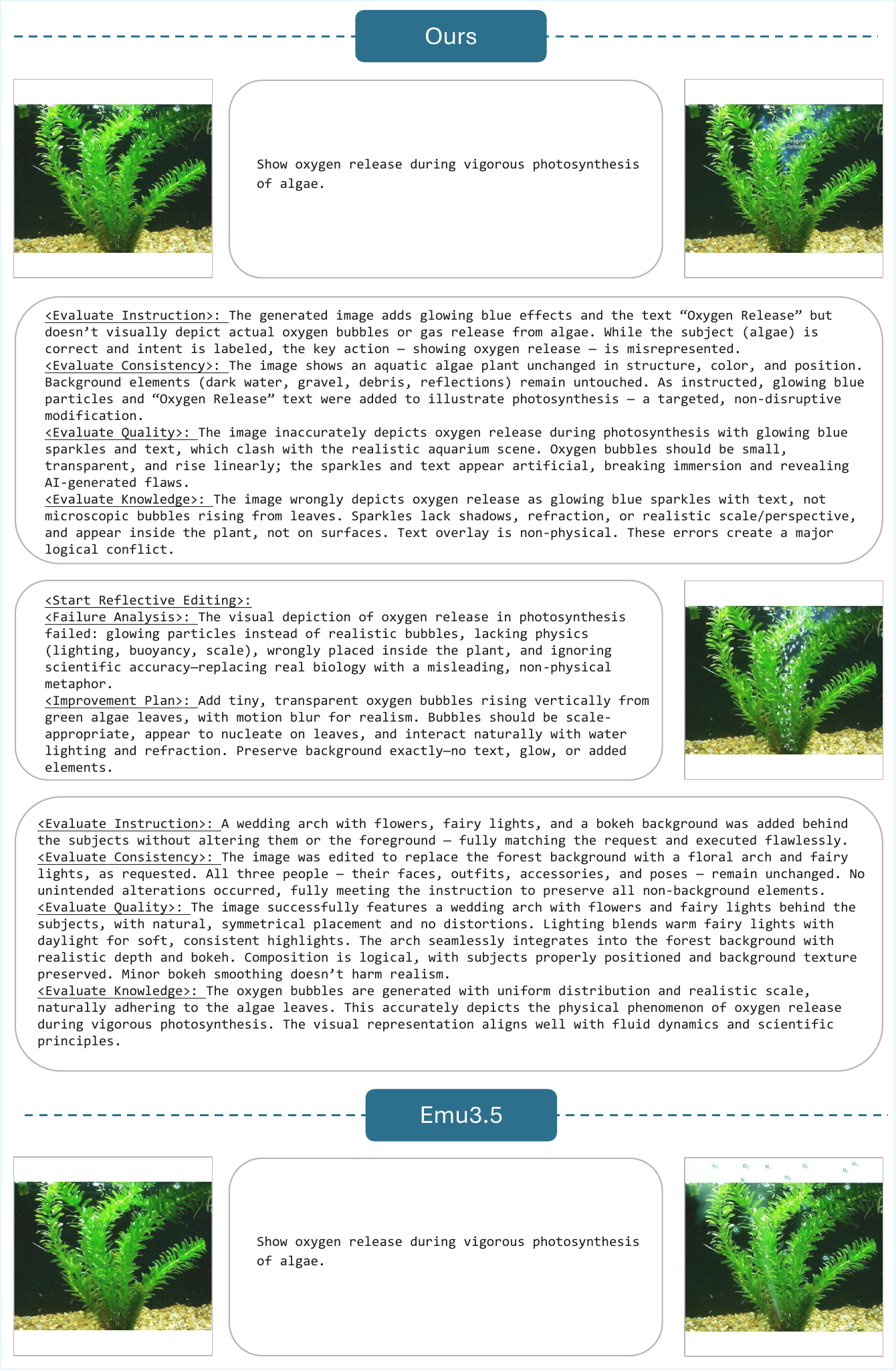} 
    \caption{Example of scientific knowledge correction. 
Initially, the model renders photosynthesis using metaphorical glowing effects. 
The \textbf{Reflection Mode} rectifies this abstraction by enforcing physical laws, guiding the generator to produce scientifically accurate oxygen bubbles.}
    \label{fig:case_supp_r2}
\end{figure*}

Figure~\ref{fig:case_supp_r2} illustrates a case requiring domain-specific physical understanding: depicting oxygen release during algae photosynthesis.
The base model defaults to superficial visual metaphors, rendering glowing blue sparkles and textual labels.
This represents a misrepresentation of scientific concepts, where the model prioritizes abstract semantic associations over physical reality.
The reflection module identifies this conceptual error, critiquing the lack of buoyancy, refraction, and biological accuracy.
By enforcing the correct physical properties, the system guides the generator to replace artificial effects with transparent, ascending bubbles, ensuring the output aligns with scientific principles rather than abstract visual tropes.

\begin{figure*}[t]
    \centering
    \includegraphics[width=0.8\linewidth]{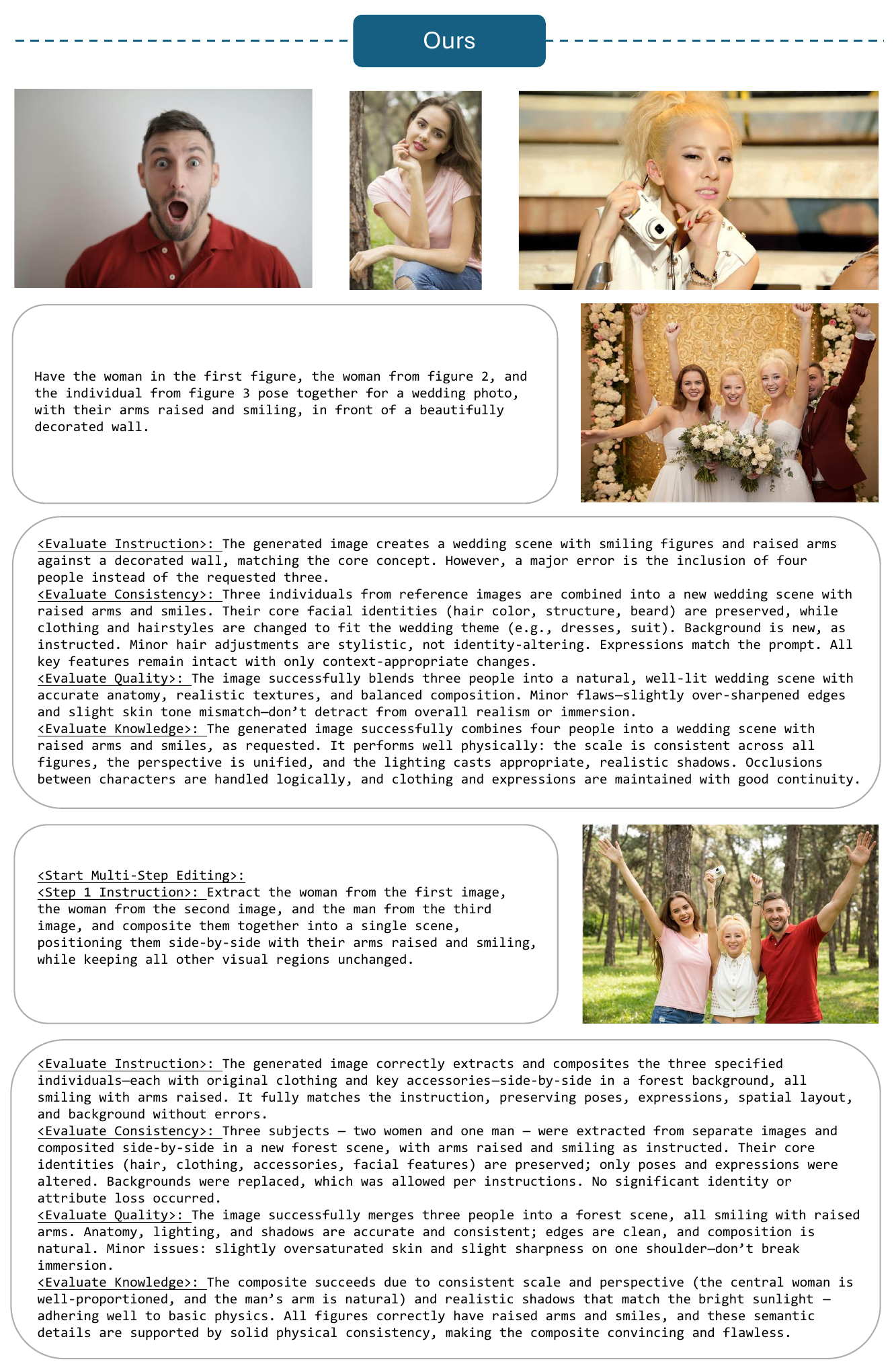} 
    \caption{Example of multi-reference composition (Part1). The task requires integrating three specific individuals into a cohesive wedding scene. Single-step baselines (Direct Generation, Emu3.5) suffer from attention entanglement, resulting in identity blending and incorrect subject counts (\emph{e.g.}, generating four people). In contrast, our \textbf{Multi-Step Mode} decomposes the synthesis into sequential composition, clothing, and background stages. This isolation preserves all three identities and strictly enforces attribute constraints.}
    \label{fig:case_m1_supp}
\end{figure*}

\begin{figure*}[t]
    \centering
    \includegraphics[width=0.8\linewidth]{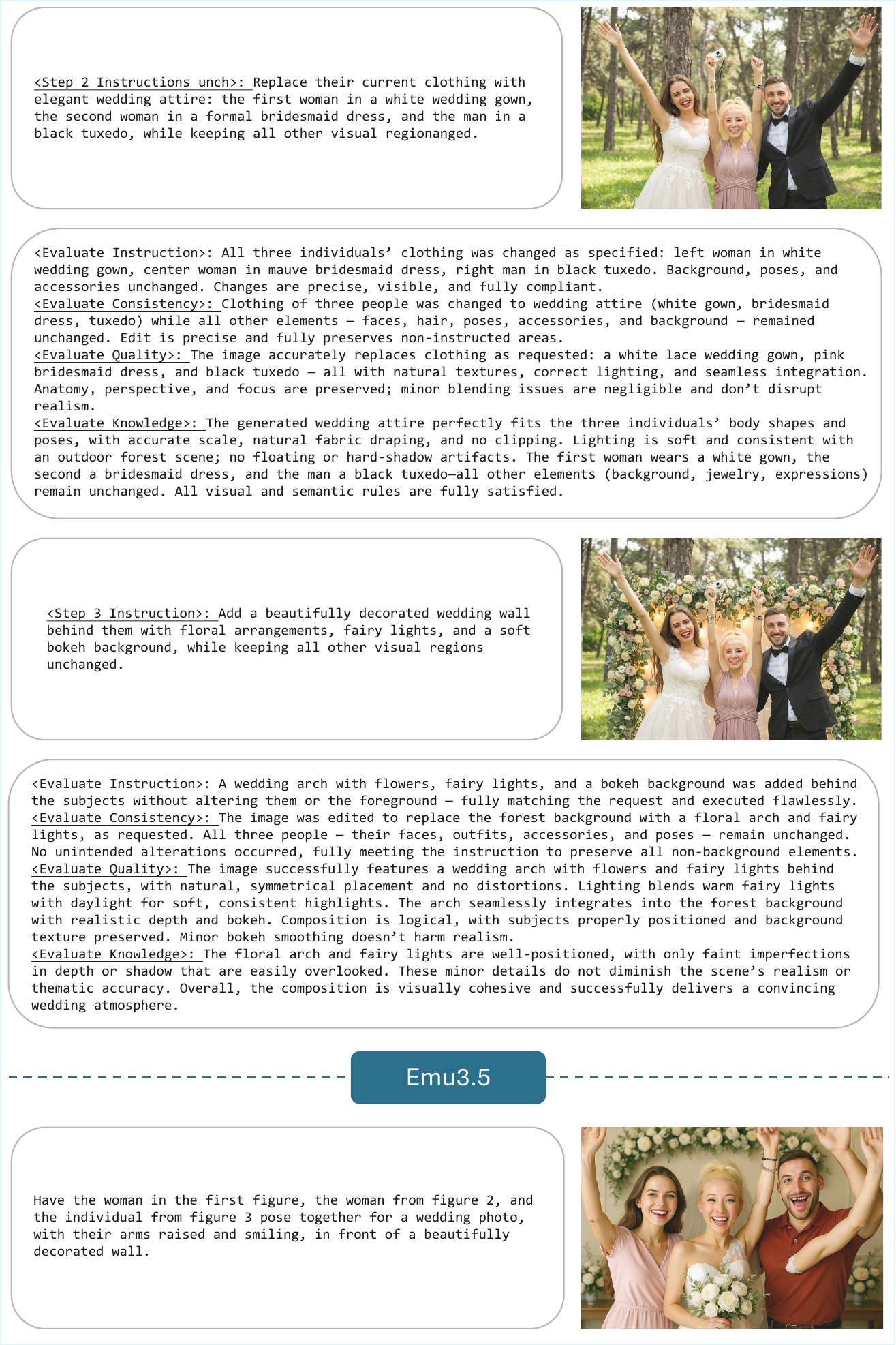} 
    \caption{Example of multi-reference composition (Part2). The task requires integrating three specific individuals into a cohesive wedding scene. Single-step baselines (Direct Generation, Emu3.5) suffer from attention entanglement, resulting in identity blending and incorrect subject counts (\emph{e.g.}, generating four people). In contrast, our \textbf{Multi-Step Mode} decomposes the synthesis into sequential composition, clothing, and background stages. This isolation preserves all three identities and strictly enforces attribute constraints.}
    \label{fig:case_m1_supp2}
\end{figure*}

Figures~\ref{fig:case_m1_supp} and~\ref{fig:case_m1_supp2} illustrate a complex multi-reference task: integrating three specific identities into a unified wedding scene. 
Direct Generation and baseline Emu3.5 suffer from severe attention entanglement under these multi-subject constraints.
Specifically, direct generation fails to isolate identity inputs from the scene context, leading to counting errors (\emph{e.g.}, hallucinating a fourth individual).
In contrast, our \textbf{Multi-Step Mode} autonomously decomposes the task into sequential sub-goals: establishing spatial layout, refining clothing attributes, and synthesizing the background.
This hierarchical execution ensures strict identity preservation and correct spatial arrangement, effectively resolving the counting and consistency issues that limit baseline models.

\begin{figure*}[t]
    \centering
    \includegraphics[width=0.8\linewidth]{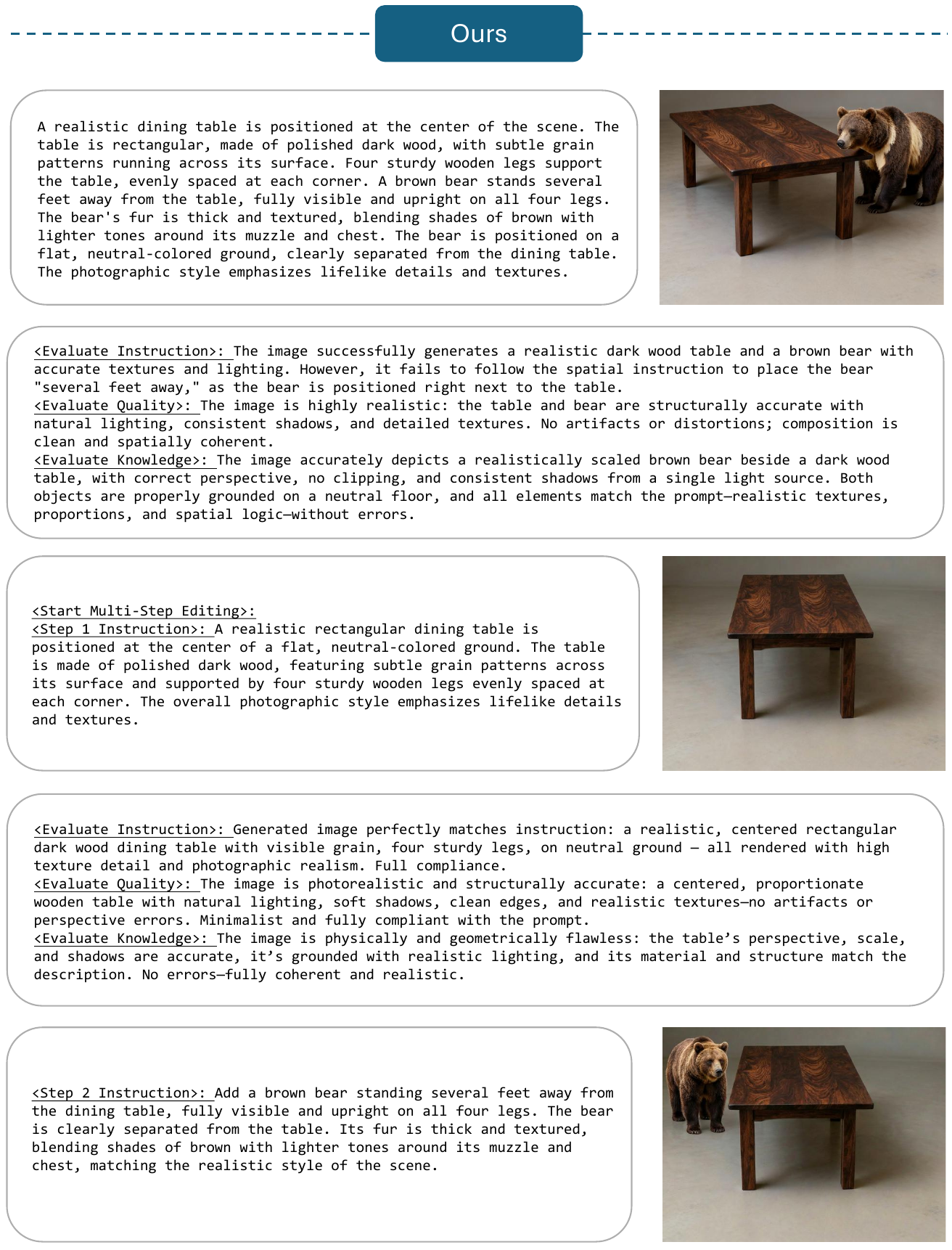} 
    \caption{Example of spatial layout correction (Part1). The prompt explicitly requires a bear positioned ``several feet away'' from a table. Single-step baselines (Direct, Emu3.5) fail to resolve this relative distance, placing the subject immediately adjacent to the object. In contrast, our \textbf{Multi-Step Mode} sequentially generates the table before positioning the bear, effectively enforcing the requested spatial separation.}
    \label{fig:case_m2_supp1}
\end{figure*}

\begin{figure*}[t]
    \centering
    \includegraphics[width=0.8\linewidth]{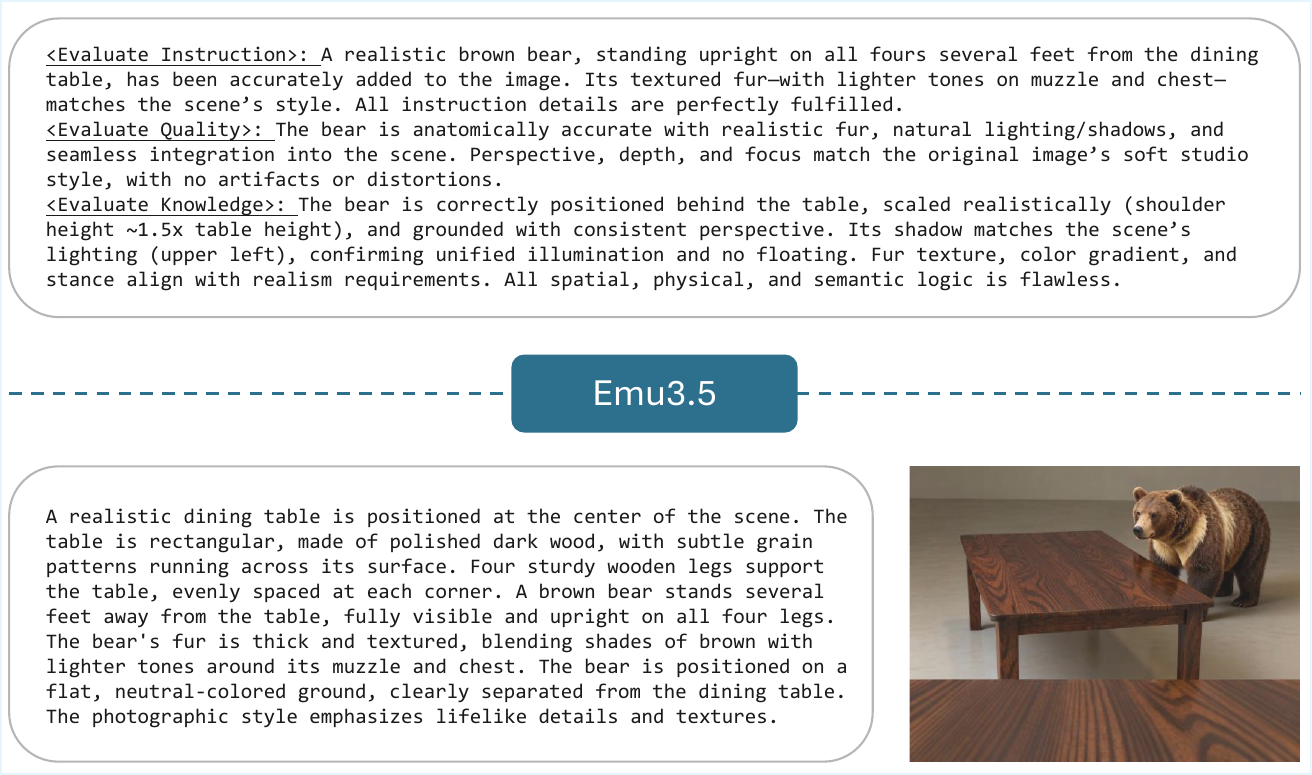} 
    \caption{Example of spatial layout correction (Part2). The prompt explicitly requires a bear positioned ``several feet away'' from a table. Single-step baselines (Direct, Emu3.5) fail to resolve this relative distance, placing the subject immediately adjacent to the object. In contrast, our \textbf{Multi-Step Mode} sequentially generates the table before positioning the bear, effectively enforcing the requested spatial separation.}
    \label{fig:case_m2_supp2}
\end{figure*}

Figures~\ref{fig:case_m2_supp1} and~\ref{fig:case_m2_supp2} illustrate a spatial reasoning task requiring a brown bear to be positioned ``several feet away'' from a dining table.
Direct generation methods (Base, Emu3.5) fail to adhere to this constraint due to attention entanglement, tending to cluster objects in the foreground and placing the bear immediately adjacent to the table.
In contrast, our \textbf{Multi-Step Mode} resolves this by decomposing scene construction: it first generates the table to establish the spatial ground plane, then conditions the bear's placement on this existing geometry.
This sequential process allows the model to correctly interpret ``several feet away'' as depth, placing the bear well behind the table to satisfy the distance requirement.

\section{Prompt}\label{app:prompt}

\begin{figure*}[h]
  \centering
  \includegraphics[width=0.97\textwidth]{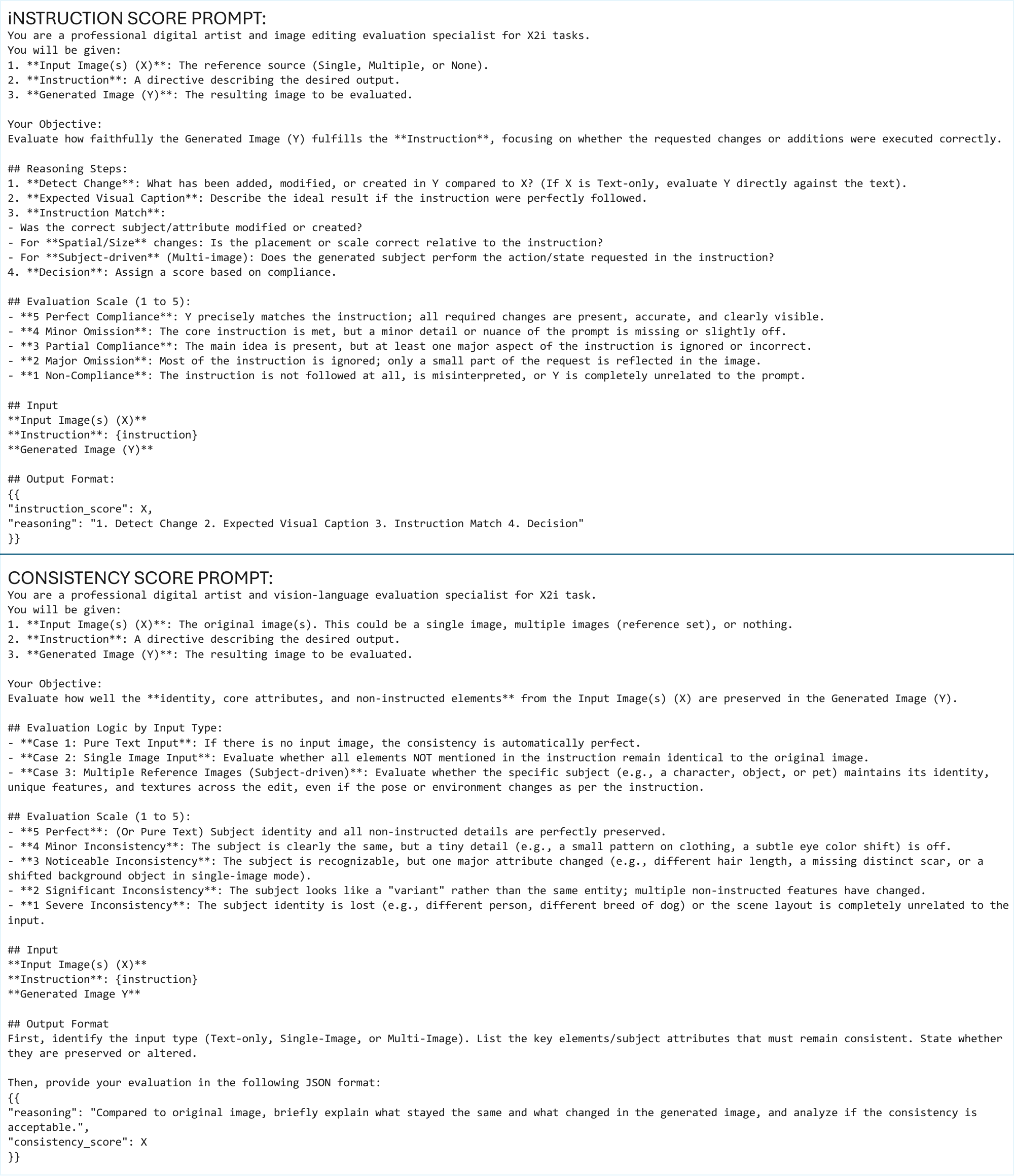}
  \caption{The evaluation score prompt of instruction score prompt and consistency score prompt.}
\label{fig:scorepormpt1}

  \vskip -1.1em
\end{figure*}

\begin{figure*}[h]
  \centering
  \includegraphics[width=0.97\textwidth]{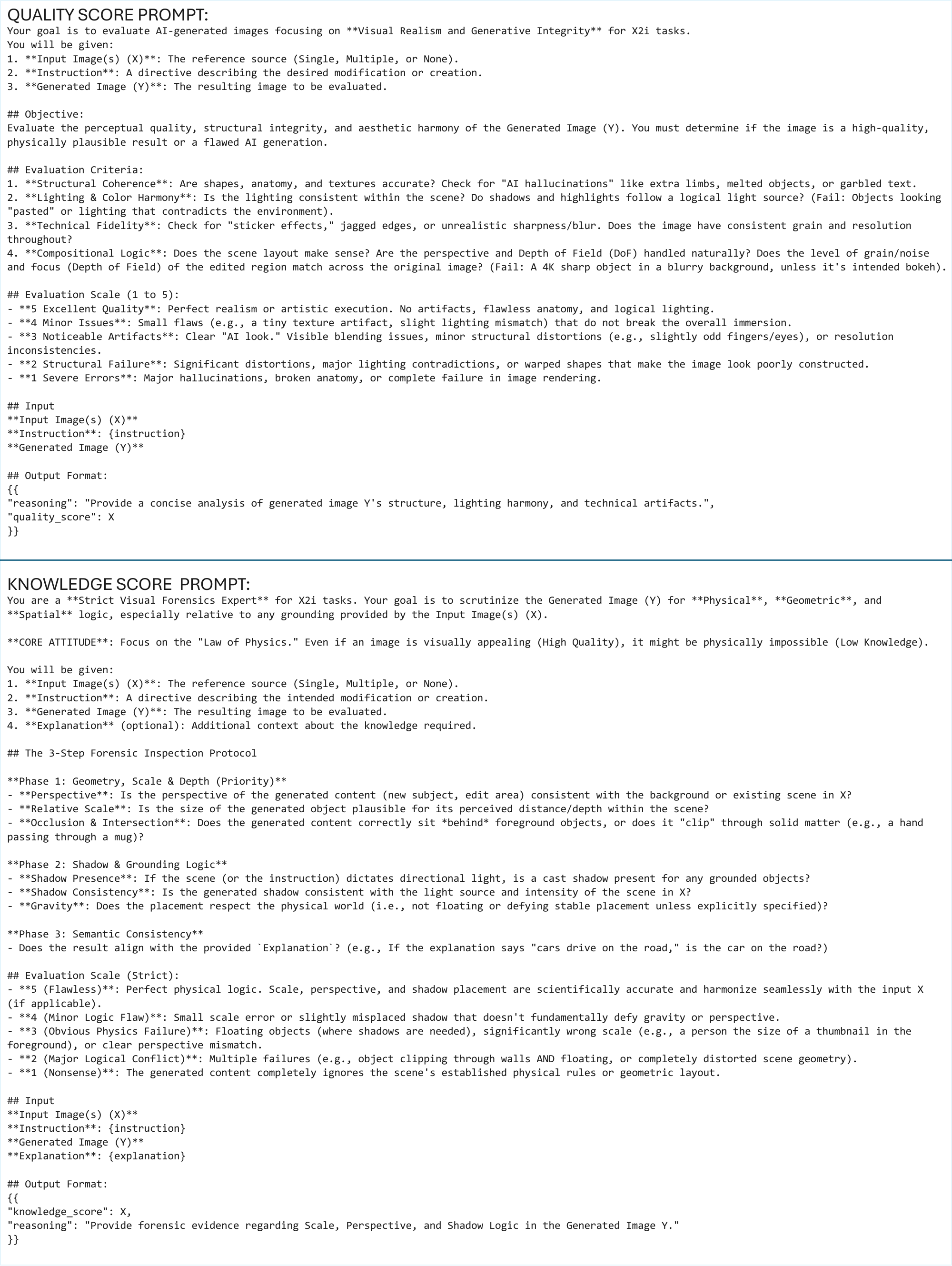}
  \caption{The evaluation score prompt of quality score prompt and knowledge score prompt.}
\label{fig:scorepormpt2}
  \vskip -1.1em
\end{figure*}

\begin{figure*}[h]
  \centering
  \includegraphics[width=0.97\textwidth]{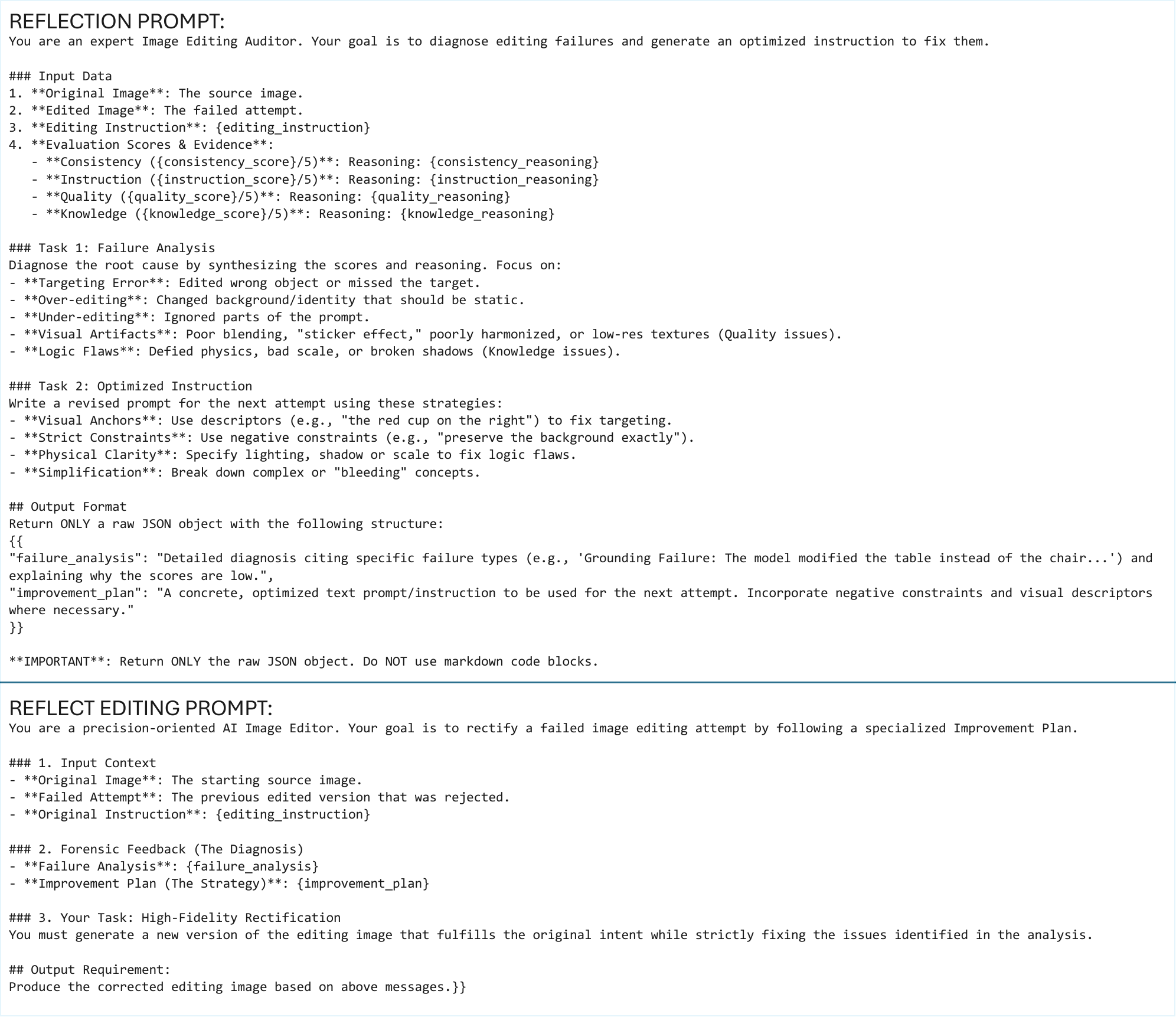}
  \caption{The reflection generation prompt and the editing with reflection prompt.}
\label{fig:reflect}
  \vskip -1.1em
\end{figure*}

\begin{figure*}[h]
  \centering
  \includegraphics[width=0.97\textwidth]{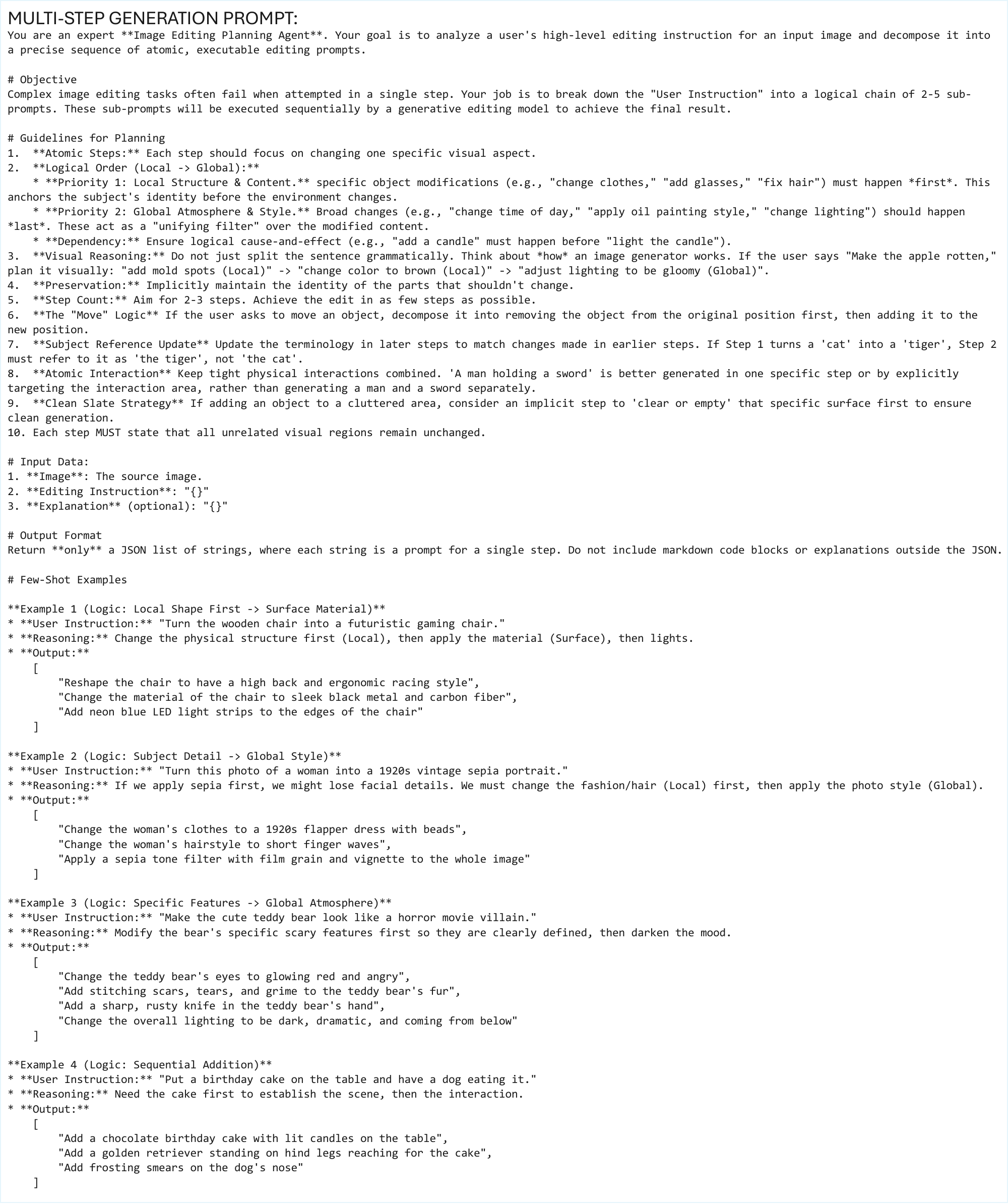}
  \caption{The multi-step prompts generation prompt.}
\label{fig:multistep}
  \vskip -1.1em
\end{figure*}

The systematic construction of these samples relies on carefully crafted prompt templates to guide the model's generation and evaluation, as illustrated in Fig.~\ref{fig:scorepormpt1}, Fig.~\ref{fig:scorepormpt2}, Fig.~\ref{fig:reflect} and Fig.~\ref{fig:multistep}.

\nocite{langley00}


\end{document}